\documentclass[onefignum,onetabnum]{siamonline250211}

\usepackage{lipsum}
\usepackage{graphicx}
\usepackage{epstopdf}

\usepackage{amsmath}
\usepackage{algorithm}
\usepackage{array}
\usepackage{textcomp}
\usepackage{stfloats}
\usepackage{url}
\usepackage{verbatim}
\usepackage{cite}
\usepackage{tikz}
\usetikzlibrary{shapes, arrows.meta, positioning}
\usepackage{amssymb}
\usepackage{mathtools}
\usepackage{bm}
\usepackage{multirow}
\usepackage{colortbl}
\usepackage{blindtext}
\usepackage{microtype}
\usepackage{booktabs}
\usepackage{algpseudocode}
\usepackage{wrapfig}
\usepackage{caption}
\usepackage{subcaption}

\newcommand{\softmax}{\mathrm{softmax}}

\ifpdf
  \DeclareGraphicsExtensions{.eps,.pdf,.png,.jpg}
\else
  \DeclareGraphicsExtensions{.eps}
\fi

\usepackage{enumitem}
\setlist[enumerate]{leftmargin=.5in}
\setlist[itemize]{leftmargin=.5in}

\newsiamremark{remark}{Remark}
\newsiamremark{hypothesis}{Hypothesis}
\crefname{hypothesis}{Hypothesis}{Hypotheses}
\newsiamthm{claim}{Claim}
\newsiamremark{fact}{Fact}
\crefname{fact}{Fact}{Facts}

\newtheorem{thm}{Theorem}[section]
\newtheorem{crl}{Corollary}[section]
\newtheorem{prop}{Proposition}[section]
\newtheorem{dfn}{Definition}[section]
\newtheorem{rmk}{Remark}[section]

\theoremstyle{remark}

\headers{Transformers with Insights from Image Filtering and Boosting}{L. Abdullaev, M. Tkachenko, and T. M. Nguyen}

\title{Revisiting Transformers with Insights from Image Filtering and Boosting}

\author{Laziz Abdullaev\thanks{Department of Mathematics, National University of Singapore 
  (\email{laziz.abdullaev@u.nus.edu}). The work is partially done during internship at Rakuten Group, Inc.}
\and Maksim Tkachenko\thanks{Frontier Research Department, Rakuten Group, Inc. 
  (\email{maksim.tkatchenko@gmail.com}).}
\and Tan M. Nguyen\thanks{Department of Mathematics, National University of Singapore
    (\email{tanmn@nus.edu.sg}).}
}
\usepackage{amsopn}

\ifpdf
\hypersetup{
  pdftitle={Aen Example Article},
  pdfauthor={L. Abdullaev, M. Tkachenko, T. M. Nguyen}
}
\fi

\begin{document}

\maketitle

\begin{abstract}
The self-attention mechanism, a cornerstone of Transformer-based state-of-the-art deep learning architectures, is largely heuristic-driven and fundamentally challenging to interpret. Establishing a robust theoretical foundation to explain its remarkable success and limitations has therefore become an increasingly prominent focus in recent research. Some notable directions have explored understanding self-attention through the lens of image denoising and nonparametric regression. While promising, existing frameworks still lack a deeper mechanistic interpretation of various architectural components that enhance self-attention, both in its original formulation \cite{vaswani2017attention} and subsequent variants. In this work, we aim to advance this understanding by developing a unifying image processing framework, capable of explaining not only the self-attention computation itself but also the role of components such as positional encoding and residual connections, including numerous later variants. We also pinpoint potential distinctions between the two concepts building upon our framework, and make effort to close this gap. We introduce two independent architectural modifications within transformers. While our primary objective is interpretability, we empirically observe that image processing-inspired modifications can also lead to notably improved accuracy and robustness against data contamination and adversaries across language and vision tasks as well as better long sequence understanding.
\end{abstract}



\section{Introduction}
Transformers \cite{vaswani2017attention} have achieved state of the art performance across a wide variety of tasks in machine learning \cite{khan2022transformers, lin2022survey} and, in particular, within natural language processing \cite{al2019character, raffel2020exploring} and computer vision \cite{touvron2021training, radford2021learning}

At the heart of transformers, there is the \textit{self-attention} (SA) mechanism, which handles capturing diverse syntactic and semantic patterns by computing weighted averages of token representations within a sequence based on the similarity scores between pairs of tokens, also known as attention scores. As will be shown in the next section, high dimensional word embeddings are combined with specifically crafted \textit{positional encodings} (PE), vectors that represent position of each token in user input, before being fed into self-attention. After the self-attention computation, output is combined with inputs, referred to as \textit{residual connection} (RC), to prevent gradient vanishing during training as well as to facilitate a better information flow through the network.

The critical significance of understanding and controlling how Transformer works makes it vital to develop approachable theoretical frameworks both to explain currently available model variants as well as shed light on potential improvements that could be gained through such proof-friendly frameworks. Consequently, recent research has attempted to study Transformers, especially the self-attention mechanism, through variety of theoretical viewpoints such as kernel smoothing \cite{tsai2019transformer}, dynamical systems \cite{lu2019understanding}, control theory \cite{Nguyen2024PIDformer}, measure theory \cite{vuckovic2020mathematicaltheoryattention}, variational denoising \cite{nguyen2023mitigating}, and recently, interacting particle systems and mean-field theory \cite{geshkovski2024mathematicalperspectivetransformers, geshkovski2024dynamic, bordelon2024infinite, bruno2025emergence} to name but a few. Diversity of such perspectives reduces the friction in engagement of researchers from various backgrounds in studying Transformers.

In a complementary vein, the broader research community has already recognized the remarkable versatility of classical image processing methods \cite{milanfar2010isdenoising, milanfar2013tour, romano2017little} in designing modern neural architectures. A particularly promising direction emerges from recent advancements in denoising algorithms applied in diffusion models \cite{Ho2020denoisingdiff, peebles2022scalable}. Modern denoising diffusion probabilistic models leverage iterative denoising refinements to generate high-fidelity images, and their underlying noise-reduction principles bear strong resemblance to filtering techniques in image processing. Moreover, nonlocal image processing techniques \cite{buades2005nlm, gilboa2007nonlocal} have found renewed relevance in deep learning. The fundamental principles behind these algorithms, including self-similarity, adaptive weighting, and spatially aware feature selection, bear strong conceptual parallels to the design of attention mechanisms \cite{vaswani2017attention, velickovic2018graph}. As a result, they offer valuable inspiration for refining the expressivity and efficiency of Transformer-based architectures in particular.

\paragraph{Contribution} In this work, we develop another theoretical perspective for Transformers that supports the understanding of the self-attention mechanism \textit{as well as} positional encoding and residual connections through the lens of image processing algorithms, which some of the related prior works were lacking \cite{nguyen2023mitigating, abdullaev2025transformer}. 

In particular, our contribution can be summarized as follows:
\begin{enumerate}
    \item We uncover a concrete connection between self-attention computation and image filtering and how an attention mechanism with properly utilized positional information could serve to make the filter \textit{bilateral} (Section \ref{section:framework}).
    \item We provide a specific, but natural, interpretation for residual connections via \textit{boosting} of iterative image filtering algorithms (particularly by increasing signal-to-noise ratio (Proposition \ref{prop:res-connection-boost})). 
    \item We theoretically prove that, with a bilateral filter-based incorporation of positional information (concatenated absolute PE or relative PE), long sequence stability of self-attention can be enhanced (Theorem \ref{thm:softmax-unstability}), which aligns with the empirical success of relative positional encoding over their absolute additive counterparts \cite{press2022train} (Remark \ref{rmk:relative-pe}).
    \item We show how a more sophisticated boosting technique can help to design residual connections to alleviate \textit{salient information} flow through deep transformers (Section \ref{section:GRC}), and offer notable robustness improvements (Theorem \ref{thm:robustness}).
\end{enumerate}

\paragraph{Organization} The paper is ordered as follows: In Section \ref{section:prelim}, we provide a brief overview of background knowledge necessary for the topic. In Section \ref{section:framework}, we present our image filtering framework for Transformers by establishing a theoretical link between its components and image processing algorithms. In Section \ref{section:GRC}, we propose a novel residual connection that is intended to improve robustness by introducing input fidelity to the output of each layer, and consequently enhance robustness. In Section \ref{section:experiments}, we provide experimental and empirical support for our theoretical observations and claims.

\section{Preliminaries}\label{section:prelim}

\subsection{Notations and Basic Definitions}

Throughout this paper, we use the following notations. Lowercase Latin letters (e.g., \( a, b, c \)) represent scalar values. Bold lowercase Latin letters (e.g., \( \bm{a}, \bm{b} \)) denote vectors. Unless stated otherwise, the coordinates of a vector \( \bm{a} \in \mathbb{R}^n \) are denoted by \( \bm{a}_i = a_i \) for $i \in [n]$, where $[n] := \{1,2, \dots, n\}$. Bold uppercase Latin letters (e.g., \( \bm{A}, \bm{B} \)) are used for matrices. The rows of a matrix \( \bm{A} \) are denoted by \( \bm{A}_i = \bm{a}_i \).



\begin{dfn}[Inequality up to a Constant Factor]
The symbol \( \lesssim \) denotes an inequality that holds up to a constant factor, i.e.,
\begin{align}
  f(n) \lesssim g(n) \iff f(n) = O(g(n)).
\end{align}
\end{dfn}

\begin{dfn}[Asymptotic Equivalence]
We use \( \asymp \) to denote asymptotic equivalence, meaning there exist positive constants \( c_1, c_2 > 0 \) such that for sufficiently large \( n \),
\begin{align}
  f(n) \asymp g(n) \iff c_1 g(n) \leq f(n) \leq c_2 g(n).
\end{align}
\end{dfn}

\subsection{Self-Attention Mechanism} \label{subsec: self-attention mechanism}

Transformers rely on multiple layers of self-attention to capture dependencies between tokens in an input sequence. Each Transformer layer consists of a self-attention block followed by feedforward layers and residual connections. The self-attention mechanism is responsible for dynamically aggregating contextual information by attending to relevant tokens, making it a crucial component for learning expressive representations. In this section, we formally define the self-attention computation and its role within the network.

\paragraph{Output computation} Given an input sequence $\bm{X} = [\bm x_1, \dots, \bm x_N]^\top \in \mathbb{R}^{N \times d}$ of $N$ feature vectors, the self-attention mechanism transforms the input to $\bm U := [\bm u_1, \dots, \bm u_N ]^\top \in \mathbb{R}^{N \times d}$ as follows:

\begin{align} \label{eqn: self-attention}
    \bm u_i &= \sum_{j = 1}^N \mathrm{softmax} \left( \frac{(\bm{W}_K\bm x_i)^{\top}(\bm{W}_Q \bm x_j)}{\sqrt{d}} \right) \bm{W}_V\bm x_j \nonumber\\
    &= \sum_{j = 1}^N \mathrm{softmax} \left( \frac{\bm q_i^\top \bm k_j}{\sqrt{d}} \right) \bm v_j
\end{align}
for $i = 1, \dots, N$, where $\softmax(a_{j}) := \softmax(\bm a)_j$ for $\bm a = [a_1,\dots,a_N]$ is an abuse of notation for convenience. The vectors $\bm q_i, \bm k_j$, and $\bm{v}_j$, $j=1,\dots,N$, are the query, key, and value vectors, respectively. They are computed as 
\begin{align}
    \bm Q &:=[\bm q_1, \dots , \bm q_N]^\top =  \bm X\bm W_Q^\top \in \mathbb{R}^{N \times d}, \nonumber\\ 
    \bm K &:= [\bm k_1, \dots , \bm k_N]^\top = \bm X\bm W_K^\top \in \mathbb{R}^{N \times d}, \nonumber\\ 
    \bm V &:= [\bm v_1, \dots , \bm v_N]^\top = \bm X\bm W_V^\top \in \mathbb{R}^{N \times d}, \nonumber
\end{align}
where $\bm W_Q, \bm W_K, \bm W_V \in \mathbb{R}^{d \times d}$ are the corresponding learnable projection matrices. Eqn.~\ref{eqn: self-attention} can then be expressed in matrix form as:
\begin{align} \label{eqn: self-attention matrix}
\bm{U} = \mathrm{softmax} \left( \frac{\bm Q\bm K^\top}{\sqrt{d}} \right) \bm V,
\end{align}
where the softmax function is applied row-wise to the matrix $\bm Q\bm K^\top / \sqrt{d}$. We refer to transformers built with Eqn.~\ref{eqn: self-attention matrix} as standard transformers or just transformers.

\paragraph{Positional Encoding} As can be seen in Eqn.~\ref{eqn: self-attention}, self-attention itself is position agnostic (e.g. it cannot account for positions of words in a sentence). For simplicity, we focus on the positional encoding of \cite{vaswani2017attention}. Given word embeddings $\bm E = [\bm e_1, \dots, e_{N}]^\top \in \mathbb{R}^{N\times d}$, \cite{vaswani2017attention} have proposed to create positional vectors $\bm{p}_i$ using the sinusoidal function:
\begin{align}\label{eq:pos-encoding}
    \begin{cases}
        \bm p_{i,2t} = \sin\left(i / T^{2t / d}\right) \\
        \bm p_{i,2t+1} = \cos\left(i / T^{2t / d}\right),
    \end{cases}
\end{align}
in which $\bm p_{i,2t}$ is the $2t^\text{th}$ coordinate of the $d$-dimensional vector $\bm{p}_i$, and the parameter $T$ is originally set to 10000. Transformer first incorporates positional information to the word embeddings by simply adding them together as
\begin{align}\label{eq:word-plus-pos}
    \bm x_i = \bm e_i + \bm p_i, \quad\quad \forall i \in [N],
\end{align}
which is then fed into the self-attention block as input as given by Eqn.~\ref{eqn: self-attention}. While the heuristic-driven design in Eqn.~\ref{eq:word-plus-pos} has proven successful in practice, we shall show how it might actually be an overshoot towards leveraging positional information in later sections.

\subsection{Data-dependent Image Filters}

The resemblance of Eqn.~\ref{eqn: self-attention} to well-known data-dependent image filtering algorithms is not difficult to observe intuitively. Before delving into formal details, we provide a brief overview of two commonly adopted classes of data-dependent image filters in this section.

Let us consider a measurement model for the denoising problem:
\begin{align}
    y_i = u_i + \eta_i, \quad\quad \forall i \in [n],    
\end{align}
where $u_i = u(\bm p_i)$ represents the underlying latent signal at location 
$\bm p_i = [p_{i,1}, p_{i,2}]^\top$, $y_i$ is the observed noisy measurement 
(e.g., pixel value), and $\eta_i$ is zero-mean white noise with variance $\sigma^2$. The task is to recover the complete set of latent signals, which we 
represent as a vector
\begin{align}
    \bm{u} = [u(\bm{p}_1), u(\bm{p}_2), \ldots, u(\bm{p}_n)]^\top,    \nonumber
\end{align}
from the corresponding noisy measurements
\begin{align}
    \bm{y} = \bm{u} + \bm{\eta}.
\end{align}
The signal $u(\bm{p})$ at position $\bm{p}$ is estimated using a (nonparametric) 
point estimation framework. Specifically, this involves solving a weighted 
least squares problem:
\begin{align}\label{eq:least-sq-denoising}
    \hat{u}(\bm{p}_i) = \arg \min_{u(\bm{p}_i)} \mathcal{J}_K(u(\bm p_i)),
\end{align}
where 
\begin{align}
    \mathcal{J}_K(u(\bm p_i)) = \sum_{j=1}^n [\underbrace{y_j - u(\bm{p}_i)}_{\text{noise } \bm{\eta}}]^2 K(\bm{p}_i, \bm{p}_j, y_i, y_j),
\end{align}
and the weight function $K(\cdot)$ is a symmetric and positive kernel
with respect to the indices $i$ and $j$, quantifying the ``similarity" between samples $y_i$ and $y_j$, located at positions $\bm{p}_i$ and $\bm{p}_j$, respectively. The stationary point condition for an optimal $\hat u$ is given by
\begin{align}
    \frac{\partial \mathcal{J}_K}{\partial u}\bigg|_{\hat u} = \sum_{j=1}^{n} -2[y_j - \hat u(\bm p_i)] K(\bm{p}_i, \bm{p}_j, y_i, y_j) = 0.
\end{align}
Then, for any given kernel function $K(\cdot)$, the solution of Eqn.~\ref{eq:least-sq-denoising} takes the following form:
\begin{align}\label{eq:weighted-averaging-soln}
    \hat{u}(\bm{p}_i) = \frac{\sum_{j=1}^{n} y_j K(\bm{p}_i, \bm{p}_j, y_i, y_j)}{\sum_{j=1}^{n} K(\bm{p}_i, \bm{p}_j, y_i, y_j)}.
\end{align}
Despite its simple form, the expression given by Eqn.~\ref{eq:weighted-averaging-soln} can represent a variety of image filters that operate by averaging similar pixels (or patches) when plugged an appropriate kernel function in. We provide two specific examples of such filters below.

\paragraph{Bilateral filter} One special class of data-dependent image filters represented by Eqn.~\ref{eq:weighted-averaging-soln} is the Bilateral Filter \cite{tomasi1998bf}, in which the kernel function is taken to account for spatial and photometric distances separably as:
\begin{align}\label{eq:bilateral-filter}
    K_{BF}(\bm{p}_i, \bm{p}_j, y_i, y_j) := &\exp\left(-\frac{\|\bm{p}_i - \bm{p}_j\|^2}{h_p^2}\right)\exp\left(-\frac{(y_i - y_j)^2}{h_y^2}\right),
\end{align}
where $h_p$ and $h_y$ are tunable parameters to control the contribution of geometric and photometric distances, repsectively.
\paragraph{Nonlocal Means (NLM) filter} Another important example is the NLM filter \cite{buades2005nlm}, which essentially generalizes the Bilateral filter by replacing the individual pixel-wise proximity in Eqn.~\ref{eq:bilateral-filter} with a patch-wise distance and discarding the effect of spatial distance as:
\begin{align}\label{eq:nlm-filter}
    K_{NLM}(\bm{p}_i, \bm{p}_j, \bm y_i, \bm y_j) := \exp\left(-\frac{\|\bm y_i - \bm y_j\|^2}{h_y^2}\right).
\end{align}
Patches could be vectorized to keep the distance metric consistent. This can also be formally regarded as letting $h_p \to \infty$ in the definition of $K_{BF}(\cdot)$ with patches $\bm y_i$ and $\bm y_j$ as last arguments. Although the spatial distance is ignored in theory, a bounded region around the position $\bm p_i$ is considered in practice to make the algorithm tractable \cite{milanfar2013tour}.

\section{A Unifying Image Processing Framework for Transformer}\label{section:framework}

To better isolate the contribution of self-attention, we focus on a simplified variant of the Transformer architecture that omits the intermediate MLP blocks typically stacked after each attention layer. A discussion of MLP layers and their role in the architecture is deferred to Section~\ref{sec:mlp}.

Below we proceed by presenting how exactly the self-attention mechanism is related to data-dependent image filter algorithms.

\subsection{Problem framing for Transformer}

We now consider the output matrix
$$\bm{U} = [\bm{u}_1, \bm{u}_2, \ldots, \bm{u}_N]^{\top} \in \mathbb{R}^{N \times d}$$
in self-attention as given by Eqn.~\ref{eqn: self-attention matrix}, where $\bm{u}_i := u(\bm p_i) \in \mathbb{R}^d$ is a real vector-valued function of positions. In the context of image processing, $\bm{U}$ can be considered as the desired \textit{clean image}, represented by vectorized patches $\bm{u}_i$ for $i \in [N]$. Further let $\bm{y}_i := y(\bm p_i) \in \mathbb{R}^d$ denote the intensity function of the $i^{\text{th}}$ patch of the observed \textit{noisy image}, that is,
\begin{align}\label{eq:transformer-denoising-problem}
\bm{y}_i = \bm{u}_i + \bm{\eta}_i, \quad\quad \forall i \in [N],
\end{align}
where $\bm{\eta}_i := \eta(\bm p_i) \in \mathbb{R}^d$ is an additive zero-mean white noise patch. We wish to reconstruct $\bm{u}_i$ from the observation $\bm{y}_i$ for all $i \in [N]$. With this denoising problem formulation specific to Transformer, the reconstruction of clean patches $\bm{u}_i$ is exactly equivalent to the weighted least squares problem given by Eqn.~\ref{eq:least-sq-denoising}. Therefore, we argue that the clean patches are estimated well by the following quantity:
\begin{align}
    \hat{\bm{u}}_i = \frac{\sum_{j=1}^{N} \bm{y}_j K(\bm{p}_i, \bm{p}_j, \bm{y}_i, \bm{y}_j)}{\sum_{j=1}^{N} K(\bm{p}_i, \bm{p}_j, \bm{y}_i, \bm{y}_j)},
\end{align}
given an appropriate weight function $K(\cdot)$. Thus, we attain the following theorem:
\begin{thm}[$1$-layer Transformer]\label{thm:self-attention-denoising}
    Consider a 1-layer transformer with a single attention head with identity projections $\bm W_{M : M \in \{Q,K,V\}} = \bm I$. Then, output vectors of its self-attention mechanism, as given by Eqn.~\ref{eqn: self-attention}, are weighted least squares estimates of clean patches $\bm{u}_i$, given noisy patches $\bm e_i$ as Eqn.~\ref{eq:transformer-denoising-problem} with the following similarity kernel function:
    \begin{align}
        K_{SA}(\bm{p}_i, \bm{p}_j, \bm{y}_i, \bm{y}_j) = \exp\left(\frac{(\bm{y}_i + \bm{p}_i)^\top (\bm{y}_j + \bm{p}_j)}{\sqrt{d}}\right).
    \end{align}
\end{thm}

\begin{rmk}
    In Theorem \ref{thm:self-attention-denoising}, we allow ourselves to take $\bm W_{M : M \in \{Q,K,V\}} = \bm I$ for simplicity. The more general case $\bm W_{M: M \in \{Q, K\}} \ne \bm I$ with proper constrains, however, can be related to learning a Mahalanobis distance to exploit the underlying data patterns to determine the most appropriate distance metric, which has proven its worth in robust imaging, regression, and deep learning tasks \cite{takeda2007lark, spira2007beltrami, nielsen2024elliptical}.
\end{rmk}

To understand an $n$-layer Transformer with $n$ self-attention blocks and residual connections through the lens of image processing, we first define the concept of \textit{boosting}. ``Boosting" in the image processing literature \cite{buhlmann2003boosting, osher2005iterative, milanfar2013tour, romano2015boosting} refers to using residual regularizing techniques enhancing image details and the signal-to-noise ratio (SNR) of the denoised image as defined below.
\begin{dfn}
    Let $\bm y = \bm u + \bm \eta$ be a noisy measurement of the clean signal $\bm u$ with noise $\bm \eta$. The SNR of the noisy observation $\bm y$ is then defined as $\sigma(\bm y) := \frac{\|\bm u\|}{\|\bm \eta\|}.$\label{eq:SNR}
\end{dfn}
When boosting is applied to denoising iterations, it effectively serves to maintain the amount of salient information exceedingly larger while allowing relatively smaller amount of noise induction. In fact, accommodating salient information flow in deep transformer layers is indeed necessary to avoid token uniformity, where all tokens tend to become almost indistinguishable, diminishing the representational capacity of the model \cite{dong2021pureattention}. 

Without loss of generality, we assume token vectors eventually collapse to a zero vector in Proposition \ref{prop:res-connection-boost} since we can subtract the limiting vector otherwise. In case of multiple limiting vectors, tokens are grouped by their limit points and the same logic is applied group-wise. This enables us to essentially quantify ``salient information" merely by their norms. Following the technique of \cite{romano2015boosting}, we now show that residual connection is indeed similar to a particular SNR boosting operation in a more general setting. 

\begin{prop}[Residual Connections Boost SNR]\label{prop:res-connection-boost}
    Consider an input vector $\bm y = \bm u + \bm \eta$ and the output vector $\hat{\bm y} = \hat{\bm u} + \hat{\bm \eta}$ of a self-attention block of Transformer, where $\bm u$ ($\hat{\bm u}$) and $\bm \eta$ ($\hat{\bm \eta}$) signify clean signal and noise in $\bm y$ ($\hat{\bm y}$), respectively. Assume that there exist constants $\alpha,\beta,\gamma \in [0, 1]$ with $\min(\alpha, \beta) > \gamma$ such that $\| \hat{\bm u} \| = \alpha \|\bm u \|$, $\cos(\bm u \mid \hat{\bm u}) \ge \beta$, and $\|\hat{\bm \eta}\| = \gamma \|\bm \eta\|$. Then, adding the input and output vectors together (residual connection) ensures an increasing SNR. That is,
    \begin{align}
        \frac{\sigma(\bm y + \hat{\bm y})}{\sigma(\bm{y})} \ge \frac{\sqrt{1 + 2\alpha\beta + \alpha^2}}{1 + \gamma} > 1.
    \end{align}
\end{prop}

We shall make the following important remarks to understand the assumption and the result of Proposition \ref{prop:res-connection-boost} before presenting its proof:

\begin{rmk}\label{rmk:denoising}
    The assumption that there exist constants $\alpha,\beta,\gamma \in [0, 1]$ with $\min(\alpha, \beta) > \gamma$ such that $\| \hat{\bm u} \| = \alpha \|\bm u \|$, $\cos(\bm u, \hat{\bm u}) \ge \beta$, and $\|\hat{\bm \eta}\| = \gamma \|\bm \eta\|$ is fundamental to image denoising since it esentially implies that the denoising output suppresses noise more than it reduces clean signal. For an \textit{ideal} denoiser, one should have $\alpha = \beta = 1$ and $\gamma = 0$.
\end{rmk}

\begin{rmk}\label{rmk:signal-vanish}
    Notice that while $\sigma(\bm y + \hat{\bm y}) > \sigma(\bm y)$, it is generally not true that $\sigma(\bm y + \hat{\bm y}) \ge \sigma(\hat{\bm y})$. However, even if SNR $\sigma(\hat{\bm y})$ happens to be greater than $\sigma(\bm y + \hat{\bm y})$ (e.g. when a denoising step was almost perfect), it does not guarantee that the absolute amount of clean signal is well-preserved. In fact, by assumption, $\|\hat{\bm u} \| = \alpha \|\bm u\|$ with $\alpha < 1$. Then recursively applying this $n$ times, we get $\|\tilde{\bm u}\| \approx \alpha^n \|\bm u\| \to 0$ as $n \to \infty$, where $\tilde{\bm u}$ denotes the denoising estimate at iteration $n$ without residual connection. In Section \ref{section:GRC}, we discuss ways to overcome this issue by modifying the residual connection approach.
\end{rmk}

\noindent\textit{Proof of Proposition \ref{prop:res-connection-boost}.} Using  Definition~\ref{eq:SNR}, we have
\begin{align}
    \frac{\sigma(\bm y + \hat{\bm y})^2}{\sigma(\bm{y})^2} &= \frac{\|\bm u + \hat{\bm u}\|^2}{\|\bm\eta + \hat{\bm \eta}\|^2}\frac{\|\bm \eta\|^2}{\|\bm u\|^2} \nonumber \\
    &= \frac{\|\bm u \|^2 + \|\hat{\bm u} \|^2 + 2\bm u^\top \hat{\bm{u}}}{\|\bm \eta \|^2 + \|\hat{\bm \eta} \|^2 + 2\bm \eta^\top \hat{\bm{\eta}}} \frac{\|\bm \eta\|^2}{\|\bm u\|^2} \nonumber \\
    &\ge \frac{\|\bm u \|^2 + \|\hat{\bm u} \|^2 + 2\|\bm u \| \|\hat{\bm{u}}\|\cos(\bm u \mid \hat{\bm u})}{\|\bm \eta \|^2 + \|\hat{\bm \eta} \|^2 + 2\|\bm \eta\| \|\hat{\bm{\eta}}\|} \frac{\|\bm \eta\|^2}{\|\bm u\|^2} \label{eq:CS} \\
    &\ge \frac{(1+\alpha^2)\|\bm u \|^2 + 2\alpha\beta\|\bm u \|^2}{(1+\gamma)\|\bm \eta \|^2 + 2\gamma\|\bm \eta\|^2} \frac{\|\bm \eta\|^2}{\|\bm u\|^2} \label{eq:cos-theta} \\
    &= \frac{1 + 2\alpha\beta + \alpha^2}{(1 + \gamma)^2}\frac{\|\bm u\|^2}{\|\bm \eta\|^2} \frac{\|\bm \eta\|^2}{\|\bm u\|^2} \nonumber \\
    &= \frac{1 + 2\alpha\beta + \alpha^2}{(1 + \gamma)^2} \label{eq:SNR-square},
\end{align}
where we utilized the fact that $\bm \eta^\top \hat{\bm{\eta}} \le \|\bm \eta\| \|\hat{\bm{\eta}}\|$ to derive \eqref{eq:CS}, then $\cos(\bm u \mid \hat{\bm u}) \ge \beta$ for \eqref{eq:cos-theta}. Note that, by assumption, we have $\min(\alpha, \beta) > \gamma$. Therefore, applying square roots in \eqref{eq:SNR-square} gives 
\begin{align}
    \frac{\sigma(\bm y + \hat{\bm y})}{\sigma(\bm{y})} &\ge \frac{\sqrt{1 + 2\alpha\beta + \alpha^2}}{1 + \gamma} \nonumber\\
    &> \frac{\sqrt{1 + 2\gamma + \gamma^2}}{1+\gamma} \nonumber\\
    &=  1 \nonumber
\end{align}
as desired. \qed

In the following corollary, we finally establish the whole Transformer architecture, excluding MLP blocks, as an iterative image processing algorithm.
\begin{crl}[$n$-layer Transformer]
    Consider an $n$-layer Transformer with each layer formed as in Theorem \ref{thm:self-attention-denoising} without the MLP blocks. Then, the computation flow of this transformer is equivalent to $n$-step iterative denoising with kernel $K_{SA}(\cdot)$ with boosting effect at each iteration via residual connections.
\end{crl}

\subsection{Distinctions of Self-Attention}\label{sec:nlsa}

Despite the simplicity of the statement of Corollary \ref{thm:self-attention-denoising}, it uncovers the hidden denoising task that self-attention is implicitly trying to solve. Nonetheless, there are a few potential concerns associated with the definition of $K_{SA}(\cdot)$ that we shall highlight through the following proposition:

\begin{prop}\label{cor:KSA-KBF} Assume $\|\bm y_i\| = c$ and let $h_p = h_y = (4d)^{1/4}$ in the definition of $K_{BF}(\cdot)$. Then, we have
\begin{align}\label{eq:KSA-KBF}
    K_{SA}(\bm{p}_i, \bm{p}_j, \bm{y}_i, \bm{y}_j) = \alpha_c K_{BF}(\bm{p}_i, \bm{p}_j, \bm{y}_i, \bm{y}_j) K_{BF}(\bm p_i, \bm y_j, \bm p_j, \bm y_i)
\end{align}
for a constant $\alpha_c > 0$ determined by $c$ and the feature dimension $d$.
\end{prop}
\noindent\textit{Proof.} First notice that 
\begin{align}\label{eq:pos-norm}
    \|\bm p_i \|^2 = \sum_{t=1}^{d/2} \left[\sin^2(i/T^{2t/d}) + \cos^2(i/T^{2t/d})\right] = \frac{d}{2}.
\end{align}
Then, Eqn.~\ref{eq:KSA-KBF} follows simply from the fact that
\begin{align}
        K_{SA}(\bm{p}_i, \bm{p}_j, \bm{y}_i, \bm{y}_j) &= \exp\left(\frac{(\bm{y}_i + \bm{p}_i)^\top (\bm{y}_j + \bm{p}_j)}{\sqrt{d}}\right) \nonumber \\
        &= \exp\left(\frac{\bm y_i^\top \bm y_j}{\sqrt{d}}\right) \exp\left(\frac{\bm p_i^\top \bm p_j}{\sqrt{d}}\right)\exp\left(\frac{\bm p_i^\top \bm y_j}{\sqrt{d}}\right)\exp\left(\frac{\bm y_i^\top \bm p_j}{\sqrt{d}}\right) \nonumber\\
        &= \alpha_c \exp\left(-\frac{\|\bm{y}_i - \bm{y}_j\|^2}{2\sqrt{d}}\right) \exp\left(-\frac{\|\bm{p}_i - \bm{p}_j\|^2}{2\sqrt{d}}\right)  \nonumber \\ &\quad\quad\quad \cdot \exp\left(-\frac{\|\bm{p}_i - \bm{y}_j\|^2}{2\sqrt{d}}\right) \exp\left(-\frac{\|\bm{y}_i - \bm{p}_j\|^2}{2\sqrt{d}}\right) \nonumber \\
        &= \alpha_c K_{BF}(\bm{p}_i, \bm{p}_j, \bm{y}_i, \bm{y}_j) K_{BF}(\bm p_i, \bm y_j, \bm p_j, \bm y_i),\nonumber
\end{align}
where $\alpha_c = \exp(2c^2 + d/2)$ is a factor coming from the norms $\| \bm y_i \| = c$ and Eqn.~\ref{eq:pos-norm}. \qed

\bigskip

Proposition \ref{cor:KSA-KBF} reveals at least \textit{three distinctions} between the self-attention mechanism and data-dependent image filters:
\begin{enumerate}
    \item Self-attention uses dot-product similarity instead of Euclidean distance which are equivalent iff content embedding vectors live on hyperspheres.
    \item The self-attention kernel redundantly (and somewhat surprisingly) possesses an extra factor of an arguably irrelevant token-to-position cross-similarity, $K_{BF}(\bm p_i, \bm y_j, \bm p_j, \bm y_i)$, as in Eqn.~\ref{eq:KSA-KBF}.
    \item Self-attention assigns equal control parameters $h_y = h_p = (4d)^{1/4}$ for token-to-token (analogous to photometric) and position-to-position (analogous to geometric) distances, which are conceptually independent quantities.
\end{enumerate}
Although the first point is quite a common practice in high dimensional spaces (and the impact of $\alpha_c$ is more or less neutralized when the weights are normalized and layer normalization is applied), the next two points actually hinder the interpretability of standard self-attention mechanisms. To address this, we  remove the factor $K_{BF}(\bm p_i, \bm y_j, \bm p_j, \bm y_i)$ from the self-attention kernel $K_{SA}(\cdot)$, simplifying the interaction to focus on direct token-to-token dependencies.

The resulting self-attention (SA) mechanism is then termed as \textit{Purely Bilateral Self-Attention} or simply Bilateral Self-Attention (BSA). In particular, BSA uses the following kernel function to compute the attention scores:
\begin{align}\label{eq:k-bsa}
    K_{BSA}(\bm p_i, \bm p_j, \bm y_i, \bm y_j) := \exp\left(\frac{\bm p_i^\top \bm W \bm p_j}{h_p^2} + \frac{\bm y_i^\top \bm W \bm y_j}{h_y^2}\right),
\end{align}
where $\bm W = \bm W_Q^\top \bm W_K$. In addition, we discuss an existing attention variant from prior work that disentangles spatial and token-wise distances by introducing separate scaling parameters, $h_p$ and $h_y$ \cite{ke2021rethinking}. While it has already been empirically studied, our framework provides natural theoretical justification for this form of decoupled modeling. When $h_p^2 \gg h_y^2$ (or $h_p \to \infty$ in theory), the weight values given by Eqn.~\ref{eq:k-bsa} essentially becomes virtually insensitive to the geometric closeness of the tokens; thus, we refer to such attention as \textit{Nonlocal Self-Attention} (NLSA), borrowing the terminology from the NLM filter in image processing. In practice, it would be equivalent to Transformer with no positional encoding \cite{haviv2022transformer}. Note that one could precompute the positional term in Eqn.~\ref{eq:k-bsa} for all layers to eliminate all extra computational costs during inference since they are input independent.

\begin{figure}
    \centering
    \includegraphics[width=0.8\linewidth]{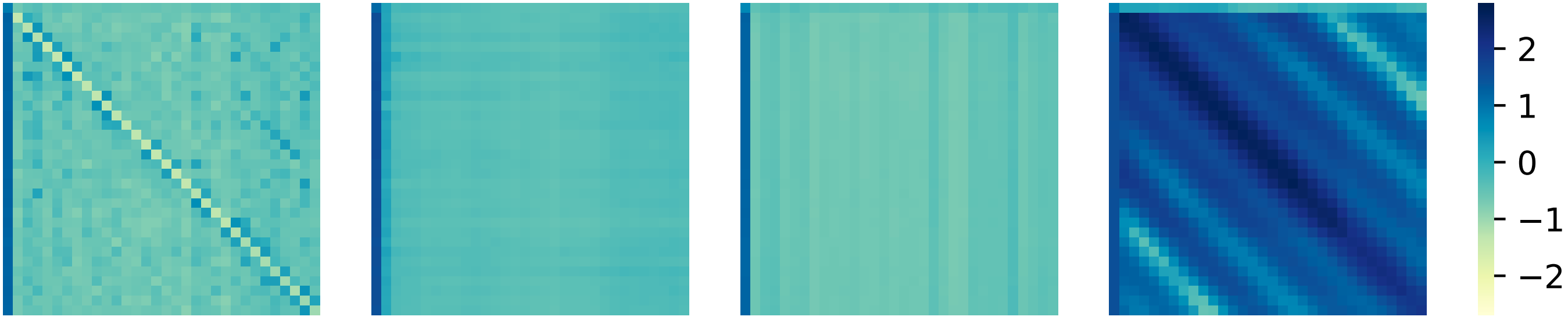}
    \caption{\textit{Left to Right}: Token-to-Token, Token-to-Position, Position-to-Token, and Position-to-Position Correlation Logit Matrices \cite{ke2021rethinking}.}
    \label{fig:enter-label}
\end{figure}

\begin{rmk}[Bilateral $\sim$ Relative PE]\label{rmk:relative-pe}
    Note that Eqn.~\ref{eq:k-bsa} can be seen as concatenating absolute PE with input tokens instead of adding, and is akin to \textit{relative PE} where the relative distance is captured by the weighted dot product of the correcponding positional vectors. In fact, any distance proxy\footnotemark{} $d_{\theta}(i, j)$ can effectively be used to implement this bilateral framework. For example, while $d_{\theta}(i,j) = \bm p_i^\top \bm W \bm p_j$ recovers the one in Eqn.~\ref{eq:k-bsa}, $d_{\theta}(i,j) = -m \left| i-j \right|$ recovers ALiBi \cite{press2022train} -- a relative PE instance proposed for a better long sequence modeling. For completeness, we compare both variants across different backbone models and various tasks in Table \ref{tab:lra}.
\end{rmk}

\footnotetext{We use the term \textit{proxy} since $d_{\theta}(i,j)$ does not have to be a proper distance metric. It is only required to decay as relative distance increases.}

The kernel function defined in Eqn.~\ref{eq:k-bsa} is similar to the kernels used by \cite{tsai2019transformer} when the weight matrix is set to be symmetric positive definite and \cite{ke2021rethinking} when positional vectors and word embeddings are equipped with independent weight matrices. The positive results reported in those works also support the idea of purely bilateral self-attention derived from our general framework.

In the following subsection, we aim to demonstrate the potential stability advantage of BSA over SA, derived from shrinking the gap between data-dependent image filters and self-attention as discussed above.

\begin{figure}
    \centering
    \includegraphics[width=\linewidth]{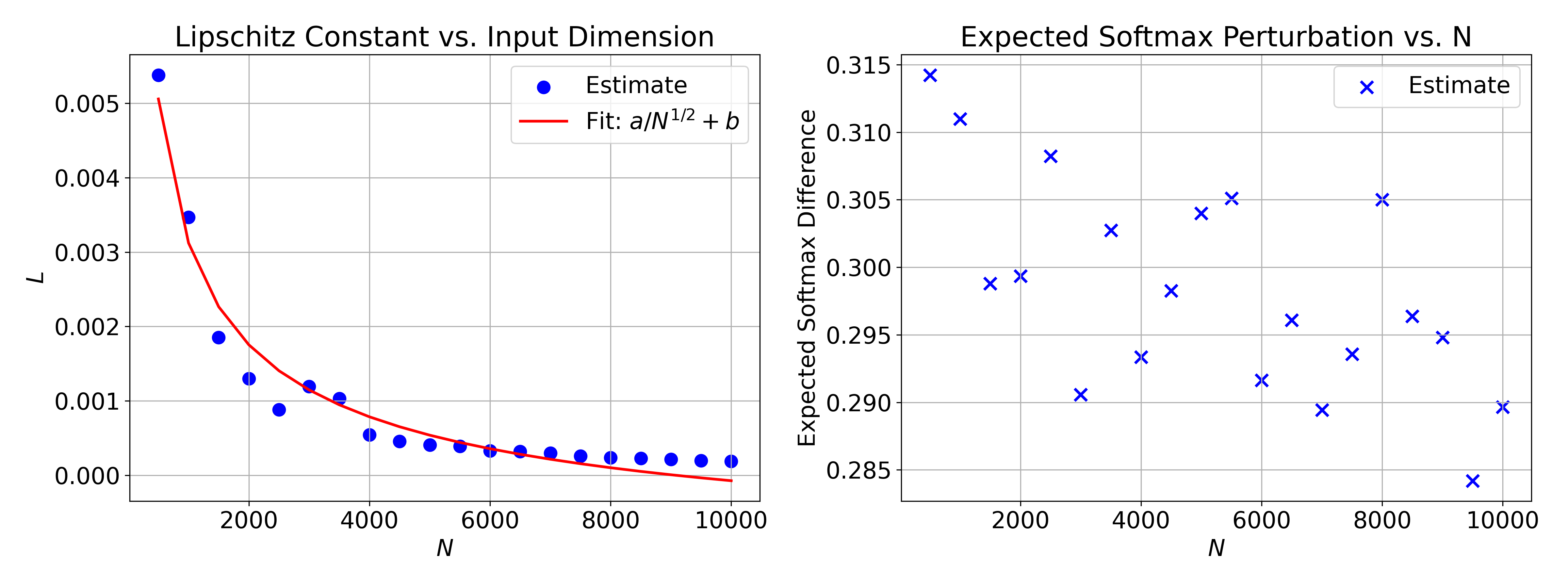}
    \caption{\textit{Left figure}: blue circles indicate the estimated tight Lipschitz constant of $\softmax$ as in Remark \ref{rmk:lipschitz} while the red curve corresponds to the best fit curve of the form $\frac{a}{\sqrt{N}}+b$. \textit{Right figure}: empirical estimate of the expectation mentioned in Theorem \ref{thm:softmax-unstability} using a random source vector $\bm c \in \mathbb{R}^N$ with $\eta_i \sim \mathcal{N}(0, 1)$, for input dimensionality $10^2 \le N \le 10^4$ using $10^3$ random samples. One can confirm that the perturbation magnitude of attention scores does not necessarily vanish with large $N$, supporting the claim of Theorem \ref{thm:softmax-unstability}.}
    \label{fig:softmax-lipschitz}
\end{figure}

\subsection{Long Range Unstability Analysis of Standard Self-Attention}\label{sec:long-range}





It has been observed empirically that token-to-position cross dot-product does not capture any particular pattern but is distributed uniformly \cite{ke2021rethinking}, effectively acting as noise. Since both token and position vectors are $d$-dimensional independent vectors with large $d$, they are close to orthogonal in expectation \cite{vershynin_high-dimensional_2018} under mild distributional assumptions. Thus, the assumption $$\eta_j := \bm p_i^\top \bm W \bm y_j + \bm y_i^\top \bm W \bm p_j \sim \mathcal{D}(0, \sigma^2)$$ is also mild, where $\mathcal{D}(0, \sigma^2)$ denotes a probability distribution (not necessarily Gaussian) with zero mean and variance $\sigma^2$. Before presenting further analysis, let us recall a useful property of the softmax function:

\begin{prop}[\cite{Gao2017OnTP}]\label{prop:softmax-lipschitz}
    The softmax function is $L$-Lipschitz with respect to the conventional Euclidean norm with $L = \lambda$, that is, for all $\bm x, \bm y \in \mathbb{R}^N$,
    \begin{align}
        \|\softmax(\bm x) - \softmax(\bm y)\| \leq \lambda \|\bm x - \bm y\|,
    \end{align}
    where $\lambda$ is the inverse temperature constant.
\end{prop}

Referencing to the Lipschitz constant of $\softmax$, in the following theorem, we derive an asymptotically tight upper bound for the perturbation magnitude in attention weights caused by containing such a noise-like summand in the input in terms of the context length $N$:

\begin{thm}[Impact on Attention Scores]\label{thm:softmax-unstability}
    Let $\bm c \in \mathbb{R}^N$ be a vector with $c_j := \bm p^\top \bm W \bm p_j + \bm e^\top \bm W \bm e_j$ for $j \in [N]$. Then, the expectation of the magnitude of softmax perturbation is asymptotically tightly bounded as follows: 
    \begin{align}
        \mathbb{E}_{\bm\eta \sim \mathcal{D}(\bm 0, \sigma^2 \bm I)}\left[\| \softmax(\bm c + \bm\eta) - \softmax(\bm c) \|\right] \le \sigma L \sqrt{N},
    \end{align}
    where $L$ is the Lipschitz constant of the softmax function.
\end{thm}

\begin{rmk}\label{rmk:lipschitz}
    Although $\softmax$ is globally Lipschitz as demonstrated in Proposition \ref{prop:softmax-lipschitz}, local behavior can be slightly more complex also depending on input dimension $N$. As depicted in Figure \ref{fig:softmax-lipschitz}, the optimal Lipschitz constant of $\softmax$, defined as
    \begin{align}
        L := \sup_{\bm x, \bm y \in S, \bm x \ne \bm y}\frac{\|\softmax(\bm x) - \softmax(\bm y)\|}{\|\bm x - \bm y\|},
    \end{align}
    \textit{can} behave (e.g. when a few components dominate) similarly to $1/\sqrt{N}$. Hence, while the upper bound in Theorem \ref{thm:softmax-unstability} seems to be unbounded, the dependence of $L$ on $N$ can make it locally $O(1)$. We can, therefore, obtain the following attainable asymptotical upper bound:
    \begin{align}
        \mathbb{E}_{\bm\eta \sim \mathcal{D}(\bm 0, \sigma^2 \bm I)}\left[\| \softmax(\bm c + \bm\eta) - \softmax(\bm c) \|\right] \lesssim \sigma.
    \end{align}
\end{rmk}

The proof of Theorem \ref{thm:softmax-unstability} is deferred to Appendix. The following immediate corollary of Theorem \ref{thm:softmax-unstability} demonstrates the effect of perturbation directly on the self-attention output vectors:

\begin{crl}[Impact on Attention Layer Output]\label{crl:err-bound-output}
    Let $\bm c \in \mathbb{R}^N$ be a vector with $c_j := \bm p^\top \bm W \bm p_j + \bm e^\top \bm W \bm e_j$ for $j \in [N]$. Then,
    \begin{align}
        \mathbb{E}_{\bm\eta \sim \mathcal{D}(\bm 0, \sigma^2 \bm I)}\left[\| \softmax(\bm c + \bm\eta)\bm V - \softmax(\bm c)\bm V \|\right] & \le \sigma L  \left\|\bm V \right\|_{\mathrm{op}}\sqrt{N} \nonumber \\&\asymp \sigma LN\sqrt{d}, \nonumber
    \end{align}
    where $\bm V \in \mathbb{R}^{N\times d}$ denotes the value matrix.
\end{crl}



Corollary \ref{crl:err-bound-output} provides the corresponding error bound on each output vector of self-attention layer since each component inside the difference norm is equivalent to one row of self-attention layer output matrix. The fact that $\| \bm V \| \asymp \sqrt{dN}$ (can be proved using the technique in the proof of Theorem \ref{thm:softmax-unstability} in Appendix) is generally an increasing quantity as $N \to \infty$ implies it contributes a notable factor to the upper bound, potentially distracting the model from learning stable representations over a sequence of considerable length. We shall provide empirical support for this claim in Section \ref{section:experiments} via long sequence modeling.

\section{Rethinking Residual Connection with Boosting}\label{section:GRC}

Although residual connections help to increase the signal magnitude at each step, they cannot completely prevent it from vanishing as mentioned in Remark \ref{rmk:signal-vanish}. To solve this issue, we will discuss leveraging older residuals aligning with a more sophisticated boosting techniques in iterative image processing.

\begin{dfn}[Generalized Residual Connection]
    Consider a transformer without MLP blocks (i.e. the output $\bm Y_{\ell}$ of layer $\ell$ is input for layer $\ell + 1$). Denote the attention computataion function at layer $\ell$ by $f_{\ell}(\cdot)$. Then, we say there is a generalized residual connection (GRC) if $\bm Y_{\ell + 1} = f_{\ell}(\bm Y_{\ell}) + t_{\ell}\bm Y_{i_{\ell}}$ where $i_{\ell}$ is sequence of indices for $\ell \in \mathbb{N}$ with $i_{\ell} \le \ell$ and $t_{\ell} \in (0, 1]$ is a scale parameter.
\end{dfn}
Note that conventional residual connection corresponds to the special case when $i_{\ell} = \ell$ and $t_{\ell} = 1$ for all $\ell \in \mathbb{N}$.

\begin{prop}\label{prop:residual-signal-vanish}
    Consider a transformer with a GRC and no MLP blocks. If there is no bounded subsequence of GRC indices $\{i_{\ell}\}_{\ell \in \mathbb{N}}$, then salient information captured by tokens asymptotically vanishes. 
\end{prop}
\textit{Proof.} Suppose the assumptions described in Remark \ref{rmk:denoising} hold. Then, the amount of salient information captured by $\bm y_{\ell+1}$ is given by 
\begin{align}
    s_{\ell} := \alpha^{\ell}\|\bm u_0\| + \alpha^{i_{\ell}}\|\bm u_0\| = \alpha^{i_{\ell}}(\alpha^{\ell - i_{\ell}} + 1)\|\bm u_0\|.
\end{align}
By definition, $\ell - i_{\ell} \ge 0$ so that $\alpha^{\ell - i_{\ell}} \le 1$ for all $\ell \in \mathbb{N}$. For the sake of contradiction, assume that all subsequences of $\{i_{\ell}\}_{\ell \in \mathbb{N}}$ are unbounded. This implies that $\lim_{\ell \to \infty} i_{\ell} = \infty$, which then yields
\begin{align}
    \lim_{\ell \to \infty} s_{\ell} = \lim_{\ell \to \infty} \alpha^{i_{\ell}}(\alpha^{\ell - i_{\ell}} + 1)\|\bm u_0\| = 0.
\end{align}
Therefore, it is necessary for the sequence $\{i_{\ell}\}_{\ell \in \mathbb{N}}$ to contain at least one bounded subsequence to avoid token uniformity. \qed

Proposition \ref{prop:residual-signal-vanish} shows that to tackle the issue of information vanishing, one ought to bound the index sequence $\{i_{\ell}\}_{\ell \in \mathbb{N}}$ for the GRC. A reasonable GRC would be to have a constant sequence $i_{\ell} = k \in \mathbb{N}$ for all $\ell > k$. To this end, we propose the boosting framework, inspired by a boosting technique for iterative image filtering \cite{romano2015boosting}. With boosting, the output of self-attention is computed as
\begin{align}\label{eq:sos-res}
\bm Y_{\ell + 1} = f_{\ell}(\bm Y_{\ell}) + t\bm Y_{0} + (1-t)\bm Y_{\ell},
\end{align}
where $t$ is a scale parameter. Rolling the Eqn.~\ref{eq:sos-res} for one more layer (without MLP and layer normalization for simplicity), the next transformer layer can be represented as
\begin{align}
    &\bm Y_{\ell + 2} = f_{\ell+1}(\bm Y_{\ell+1}) + t\bm Y_{0} + (1-t)\bm Y_{\ell+1} \nonumber \\
    &= f_{\ell+1}(f_{\ell}(\bm Y_{\ell}) + t\bm Y_{0} + (1-t)\bm Y_{\ell}) + t\bm Y_{0} + (1-t)\bm Y_{\ell+1} \nonumber \\
    &= f_{\ell+1}(\underbrace{f_{\ell}(\bm Y_{\ell}) + \bm Y_{\ell}}_{\text{original signal}} + t\underbrace{(\bm Y_{0} - \bm Y_{\ell})}_{\text{residual}}) + \underbrace{t\bm Y_{0} + (1-t)\bm Y_{\ell+1}}_{\bm R_{\ell + 1}} \nonumber \\
    &= f_{\ell+1}(f_{\ell}(\bm Y_{\ell}) + \bm Y_{\ell}) + tf_{\ell+1}(\bm Y_{0} - \bm Y_{\ell}) + \bm R_{\ell + 1}. \label{eq:twicing}
\end{align}
The boosting procedure, therefore, can essentially be seen as strengthening the signal by overlaying the residual and current signals (higher SNR due to Proposition \ref{prop:res-connection-boost}), filtering noise in a strengthened signal. Noting that the attention computation is a pseudo-linear (linear with data-dependent entries) transformation, Eqn.~\ref{eq:twicing} can recover many boosting techniques for image denoising algorithms such as twicing. Twicing, originally proposed by \cite{tukey77}, effectively re-uses the smoothed residual after each iteration step of denoising by combining it with the new estimate to get a more fine-grained output. In Figure \ref{fig:oversmoothing}, we verify that vanilla attention with the standard residual connection tends to progressively smooth the input tokens across layers, and boosting alleviates the issue.

Now to understand the potential of boosting to improve adversarial robustness, we compare Lipschitzness of networks equipped with the standard residual connection (RC) update rule
\begin{align}
    \bm{Y}_{\ell+1} = f_\ell(\bm{Y}_\ell) + \bm{Y}_\ell,
\end{align}
to the modified, generalized residual connection (GRC) as given by Eqn.~\ref{eq:sos-res}, where $f_\ell(\cdot)$ represents the layer computation (e.g., self-attention), and $t \in (0,1]$ is a parameter. The Lipschitz constant of a function quantifies its sensitivity to input changes. A function $g$ is $K$-Lipschitz if for all $\bm{x}, \bm{x}'$, $\| g(\bm{x}) - g(\bm{x}') \| \leq K \| \bm{x} - \bm{x}' \|$.
We define $K_\ell$ such that $\| \bm{Y}_\ell - \bm{Y}_\ell' \| \leq K_\ell \| \bm{Y}_0 - \bm{Y}_0' \|$, and aim to compare $K_n$, the $n$-layer Lipschitzness:

\begin{thm}[Robustness of Boosting]\label{thm:robustness}
    Consider a network with $n$ layers, where each layer $f_\ell$ has Lipschitz constant $L>0$, i.e., $\| f_\ell(\bm{u}) - f_\ell(\bm{v}) \| \leq L \| \bm{u} - \bm{v} \|$. Then, for all $t \in [0,1]$, we have
    \begin{align}
    \frac{K^{\text{GRC}}_n}{K^{\text{RC}}_n} \asymp \left( 1 - \frac{t}{L + 1} \right)^n,
    \end{align}
    implying that GRC effectively reduces the error propagation as $t \le 1 < L+1$, enhancing robustness.
\end{thm}

The proof of Theorem \ref{thm:robustness} is provided in Appendix.

\section{Experimental Results}\label{section:experiments}



\subsection{Language Modeling on WikiText-103}


We follow the experimental setup of \cite{nielsen2024elliptical} by training a small backbone that consists of 16 layers, 8 attention heads with a dimension of 16, a feedforward layer with a size of 2048, and an embedding dimension of 128. The results in Table \ref{tab:wiki} indicate a general trend of improvement in language modeling performance under both clean and contaminated data settings on the WikiText-103 dataset. Starting with the baseline \textit{Transformer} model, we observe a decrease in perplexity (PPL) when a boosting mechanism and Bilateral Attention is added. The contamination attack follows the method of random word swapping as described in \cite{merity2016pointer}.

\subsection{Long Sequence Modeling on the LRA Benchmark}

To empirically verify the propositions made in Section \ref{sec:long-range}, we experiment on long sequence modeling, adopting the setup of \cite{chen2023primal}. For each of the 5 tasks in the Long Range Arena (LRA) benchmark \cite{tay2020long}, equation calculation (ListOps), review classification (Text), document retrieval (Retrieval), image classification (Image), and image spatial dependencies (Pathfinder), we compare Bilateral Attention against standard self-attention \cite{vaswani2017attention} as well as ALiBi \cite{press2022train}. Model backbones for all three models--Transformer \cite{vaswani2017attention}, Linformer \cite{wang2020linformer}, and Nystr\"omformer \cite{xiong2021nystromformer}--are set with 2 layers, hidden dimension of 128, 2 attention heads of dimension 32, and embedding dimension of 64. Only for the case of Linformer, we interpolate between standard PE \cite{vaswani2017attention} and other two models by combining input tokens with absolute PE vectors, scaled down by 0.3, before integrating ALiBi or Bilateral Attention. This introduces controlled noise in the attention scores to mitigate rapid overfitting due to the considerable number of new learnable parameters introduced by Linformer. All other training configurations are kept the same for maximum fair comparison across the variants.

\begin{figure*}
    \centering
    \begin{subfigure}[b]{0.245\textwidth}
        \centering
        \includegraphics[width=\linewidth]{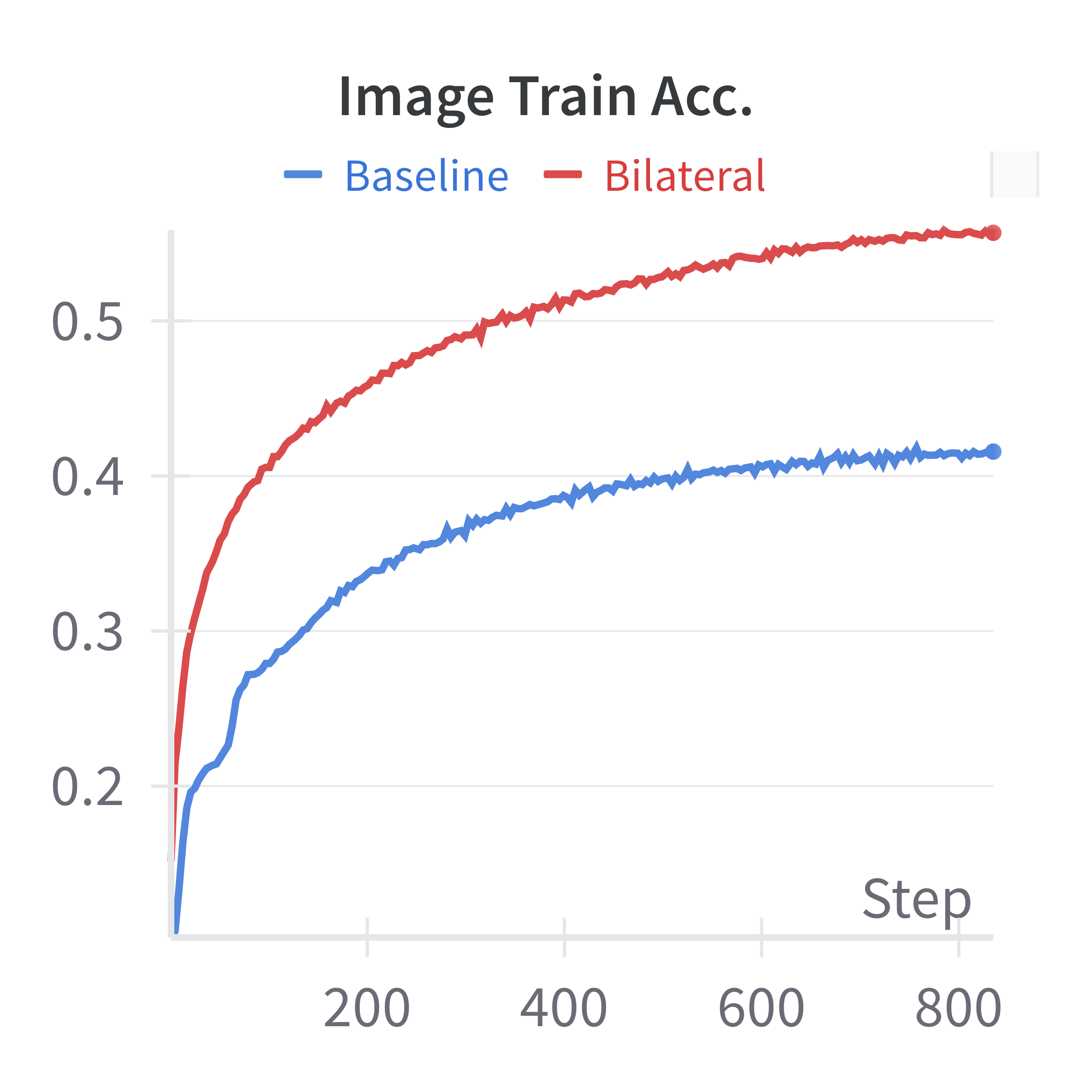}
    \end{subfigure}%
    \hfill
    \begin{subfigure}[b]{0.245\textwidth}
        \centering
        \includegraphics[width=\linewidth]{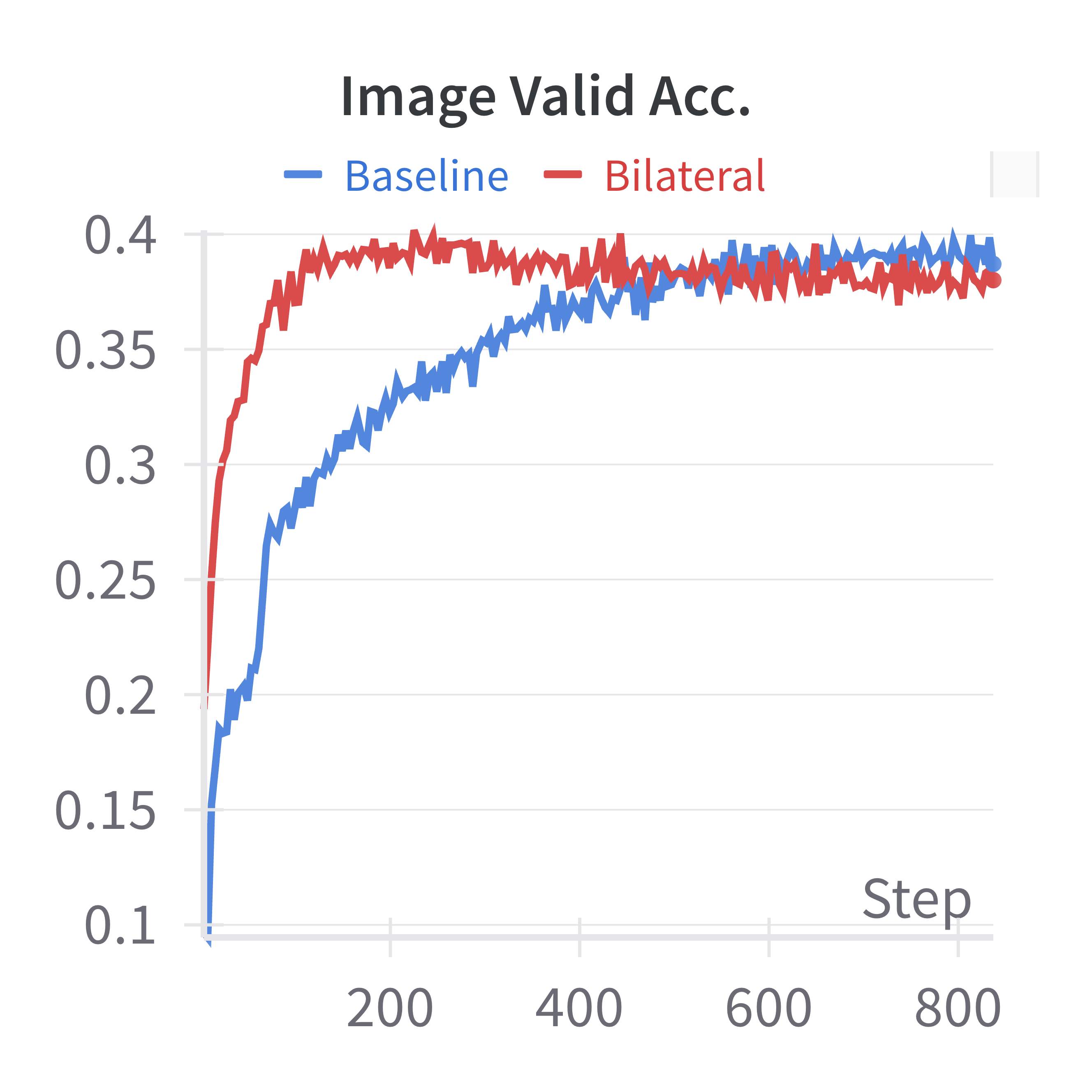}
    \end{subfigure}%
    \hfill
    \begin{subfigure}[b]{0.245\textwidth}
        \centering
        \includegraphics[width=\linewidth]{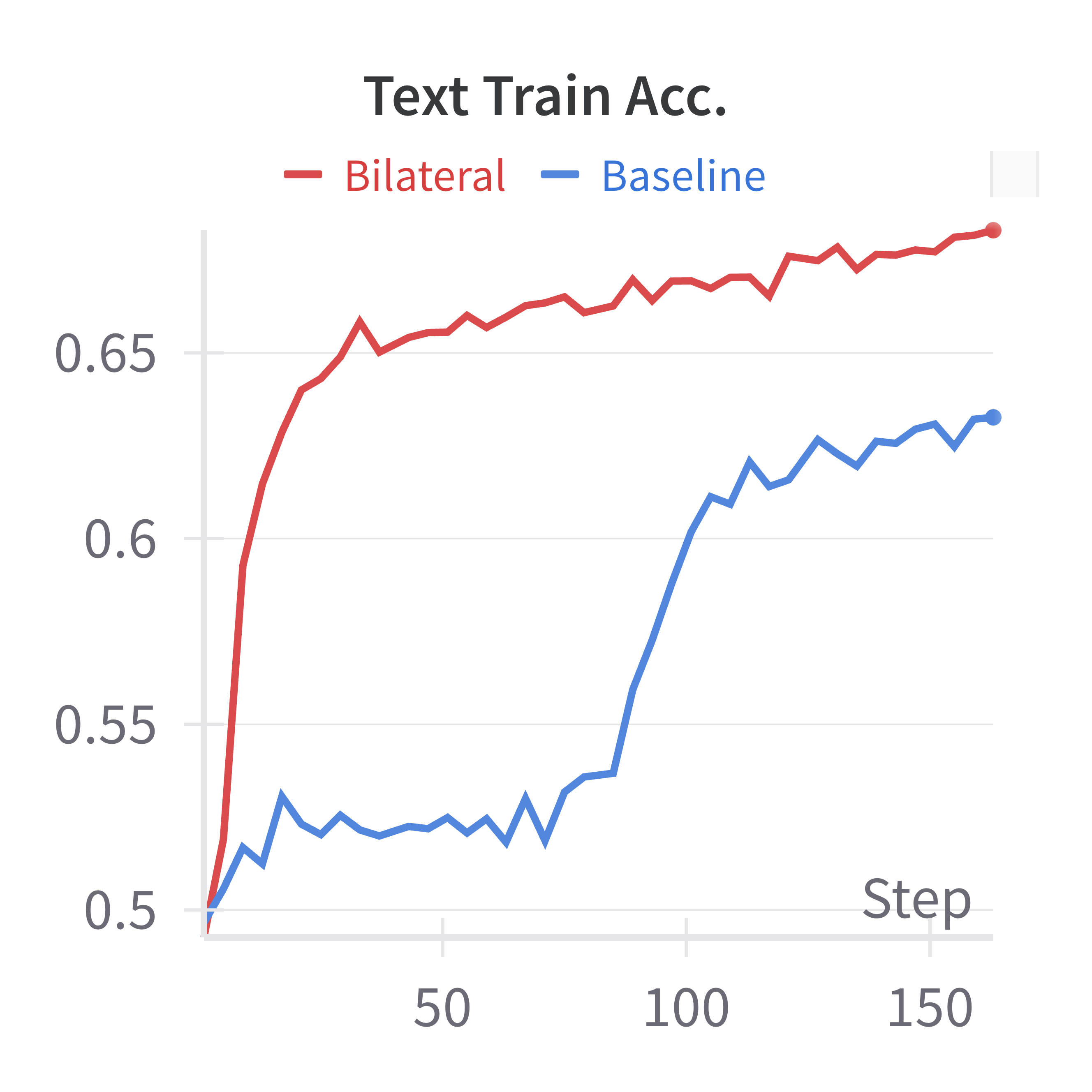}
    \end{subfigure}%
    \hfill
    \begin{subfigure}[b]{0.245\textwidth}
        \centering
        \includegraphics[width=\linewidth]{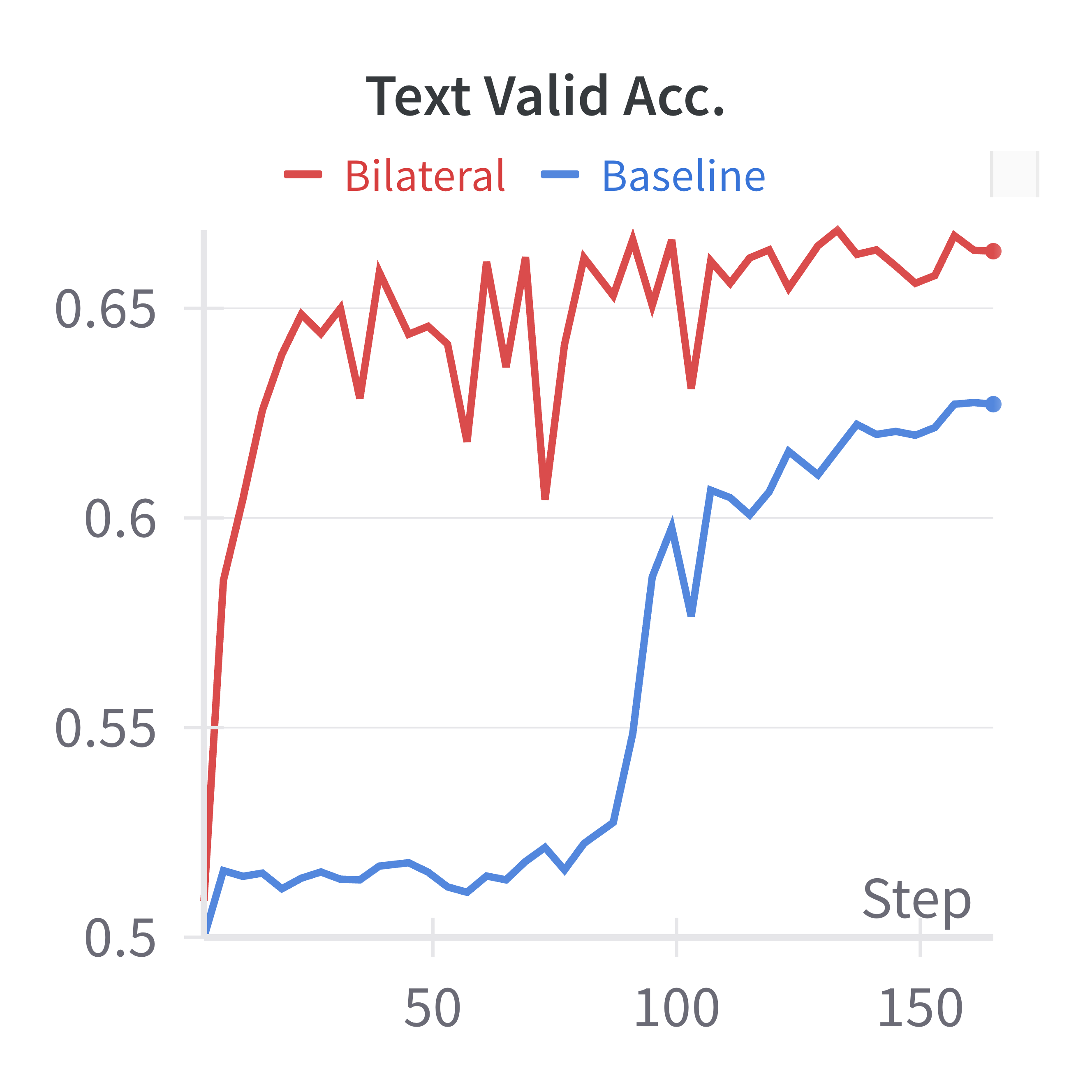}
    \end{subfigure}%
    \caption{LRA \cite{tay2020long} learning curves of \textcolor{blue}{baseline} \cite{vaswani2017attention} and \textcolor{red}{bilateral} attention mechanisms for Image (1K) and Text (4K) datasets. The figures show that the vanilla self-attention either learns substantially slower compared to its bilateral counterpart or struggles to learn useful long context relations.}
    \label{fig:data-overview}
\end{figure*}

It is observed that models with Bilateral Attention outperform baseline models and ALiBi in 4 of 5 tasks (Table \ref{tab:lra}), especially with significantly faster convergence than absolute PE during training as shown in Figure \ref{fig:data-overview}. This is particularly evident from the hardest task in the benchmark--document retrieval (Retrieval). The context length for the retrieval task is 4K and the required attention span (RAS) is around 1.3K, making it the most demanding among all reported tasks.

\begin{table}[t]
  \small
  \centering
  \caption{Test accuracy comparison on the Long Range Arena (LRA) benchmark, transposed so that methods are rows and tasks are columns.}
  \label{tab:lra}
  \begin{tabular}{lccccc}
    \toprule
    Method & Pathfinder (1K) & Image (1K) & ListOps (2K) & Text (4K) & Retrieval (4K) \\
    \midrule
    \textit{Transformer} \cite{vaswani2017attention} & 67.83 & 39.90 & 18.55 & 62.73 & 61.31 \\
    + ALiBi & 62.31 & \underline{43.42} & 18.35 & \textbf{72.95} & 74.47 \\
    \rowcolor{gray!20}+ Bilateral & 62.27 & 39.92 & \underline{37.70} & 66.85 & 76.38 \\
    \midrule
    \textit{Linformer} \cite{wang2020linformer} & 51.22 & 42.69 & 19.15 & 55.94 & 77.68 \\
    + ALiBi & 60.82 & 41.64 & 37.15 & 55.37 & \underline{78.58} \\
    \rowcolor{gray!20}+ Bilateral & 51.22 & 42.80 & \textbf{38.26} & 57.05 & \textbf{78.65} \\
    \midrule
    \textit{Nystr\"{o}mformer} \cite{xiong2021nystromformer} & \underline{68.22} & 41.23 & 18.70 & 62.16 & 60.89 \\
    + ALiBi & 61.07 & 41.27 & 20.82 & \underline{67.48} & 70.11 \\
    \rowcolor{gray!20}+ Bilateral & \textbf{69.00} & \textbf{45.56} & 37.00 & 66.18 & 73.84 \\
    \bottomrule
  \end{tabular}
\end{table}

\begin{table}[t]
  \small
  \centering
  \caption{Language Modeling on Wikitext-103}
  \label{tab:wiki}
  \begin{tabular}{lccc}
    \toprule
    Model & Test PPL ($\downarrow$) & Attacked PPL ($\downarrow$)  \\
    \midrule
    \textit{Transformer} \cite{vaswani2017attention} & 34.29 & 55.21 \\
    + Boost & {33.84} & 54.62 \\
    + Bilateral & \underline{33.40} & \underline{54.23} \\
    \rowcolor{gray!20}
    + Bilateral + Boost & \textbf{33.12} & \textbf{53.97} \\
    \bottomrule
  \end{tabular}
\end{table}

\subsection{Vision Tasks on ImageNet-1k and ADE20k}

In Table \ref{tab:imnet} and Table \ref{tab:ade20k}, we report the results on image classification on ImageNet-1k \cite{russakovsky2015imagenet} and segmentation on ADE20k \cite{zhou2019semantic}, comparing 4 models: DeiT \cite{touvron2021training} as a baseline, DeiT+Boost is our integration of GRC on top of residual connection, DeiT+Bilateral is our implementation of Bilateral Attention within the DeiT model replacing the standard self-attention, and DeiT+Bilateral+Boost is the model for integrating GRC on top of Bilateral Attention. The DeiT backbone uses 12 layers, 3 heads of dimension 64, patch size 16, feedforward layer of size 768 and embedding dimension of 192 similar to \cite{nielsen2024elliptical}, trained for 300 epochs. For DeiT+Boost, we set the parameter $t$ to learnable with initial value $0$. The experiments are carried on clean images as well as under two widely adopted adversarial attacks Fast Gradient Sign Method (FGSM) \cite{madry2017towards} and Projected Gradient Descent (PGD) \cite{goodfellow2014explaining} attacks. Additionally, Fully Attentional Network (FAN) backbone \cite{zhou2022understanding} is also tested in Table \ref{tab:imnet} together with our lightweight Boosting method to verify its compatibility with different backbones. Both FAN and and FAN+Boost models are trained for 130 epochs, adopting the training settings from \cite{nielsen2024elliptical}. 

\begin{table}[t]
  \small
  \centering
  \caption{Object Classification on ImageNet-1K without and with adversarial attack of perturbation budget 1/255.}
  \label{tab:imnet}
  \begin{tabular}{lcccccc}
    \toprule
    \multirow{2}{*}{Model} & \multicolumn{2}{c}{No Adversary} & \multicolumn{2}{c}{FGSM} & \multicolumn{2}{c}{PGD} \\
    & Top 1 & Top 5  & Top 1 & Top 5 & Top 1 & Top 5  \\
    \midrule
    \textit{DeiT} \cite{touvron2021training} & 71.97 & 90.99 & 55.15 & 85.35 & 43.87 & 78.47 \\
    + Boost & 72.22 & 91.26 & {56.10} & 85.76 & \textbf{45.27} & \textbf{79.56} \\
    + Bilateral & \underline{72.83} & \underline{91.50} & \underline{56.18} & \underline{86.02} & {43.77} & {77.98} \\
    \rowcolor{gray!20} \,$\vert$\, + Boost & \textbf{73.42} & \textbf{91.84} & \textbf{57.68} & \textbf{86.66} & \underline{44.57} & \underline{78.75} \\
    \midrule
    \textit{FAN} \cite{zhou2022understanding} & 78.85 & 94.40 & 54.12 & 83.84 & 45.24 & 80.59 \\
    \rowcolor{gray!20}+ Boost & \textbf{79.06} &\textbf{94.82} & \textbf{55.26} & \textbf{86.39} & \textbf{46.15} & \textbf{82.42} \\
    \bottomrule
  \end{tabular}
\end{table}

\begin{table}[t]
  \small
  \centering
  \caption{Image Segmentation on ADE20k}
  \label{tab:ade20k}
  \begin{tabular}{lccc}
    \toprule
    Model & Pixel Acc & Mean Acc & Mean IoU  \\
    \midrule
    \textit{DeiT} \cite{touvron2021training} & 76.12 & 40.52 & 31.09 \\
    + Boost & 76.36 & \underline{41.86} & \underline{32.02} \\
    + Bilateral & \underline{76.40} & 41.68 & 31.79 \\
    \rowcolor{gray!20}
    + Bilateral + Boost & \textbf{77.18} & \textbf{42.04} & \textbf{32.23} \\
    \bottomrule
  \end{tabular}
\end{table}

\begin{figure*}[t]
    \centering
    \begin{minipage}{0.5\linewidth}
        \centering
        \includegraphics[width=\linewidth]{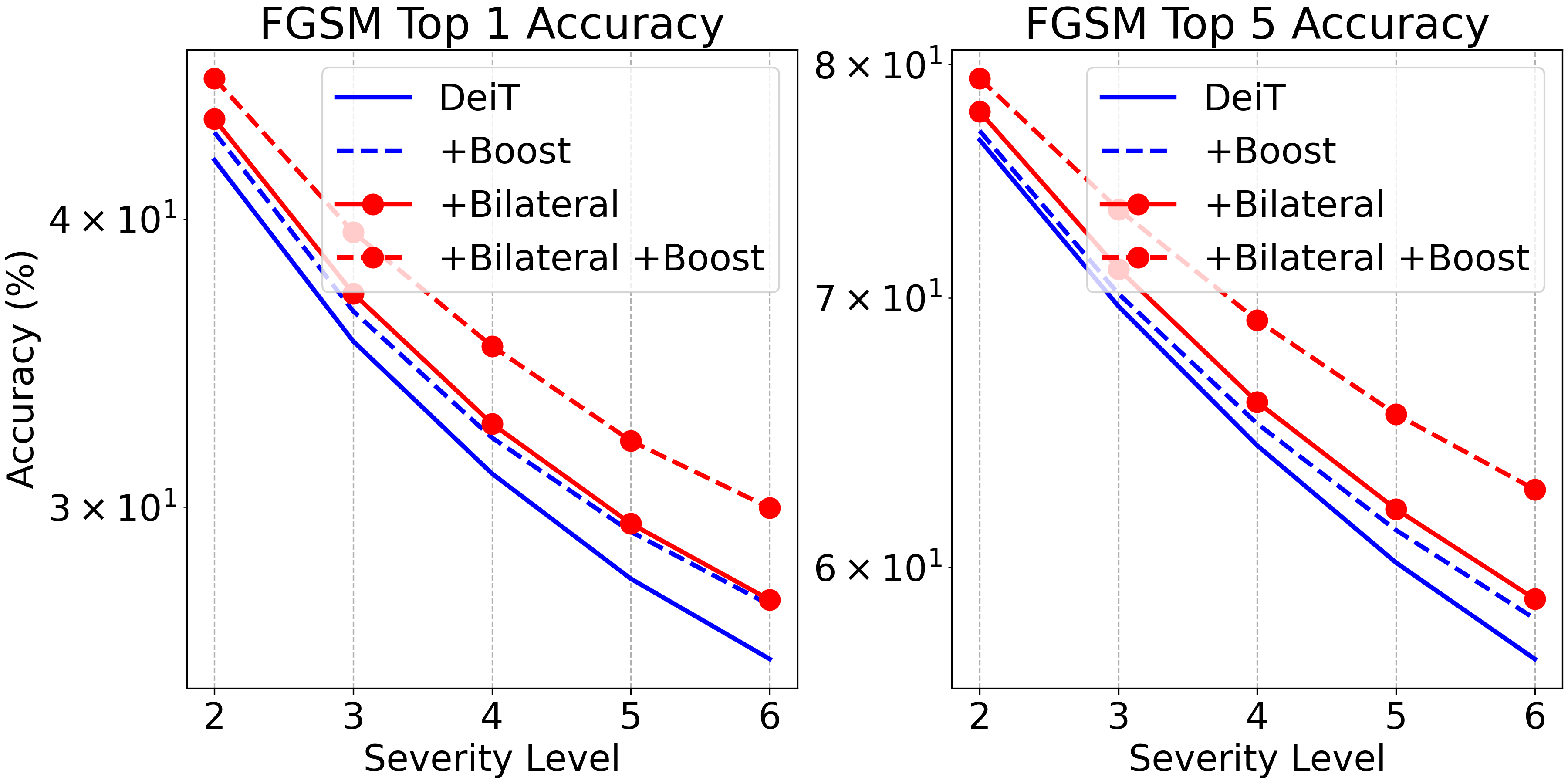}
        \label{fig:fgsm}
    \end{minipage}%
    \begin{minipage}{0.5\linewidth}
        \centering
        \includegraphics[width=\linewidth]{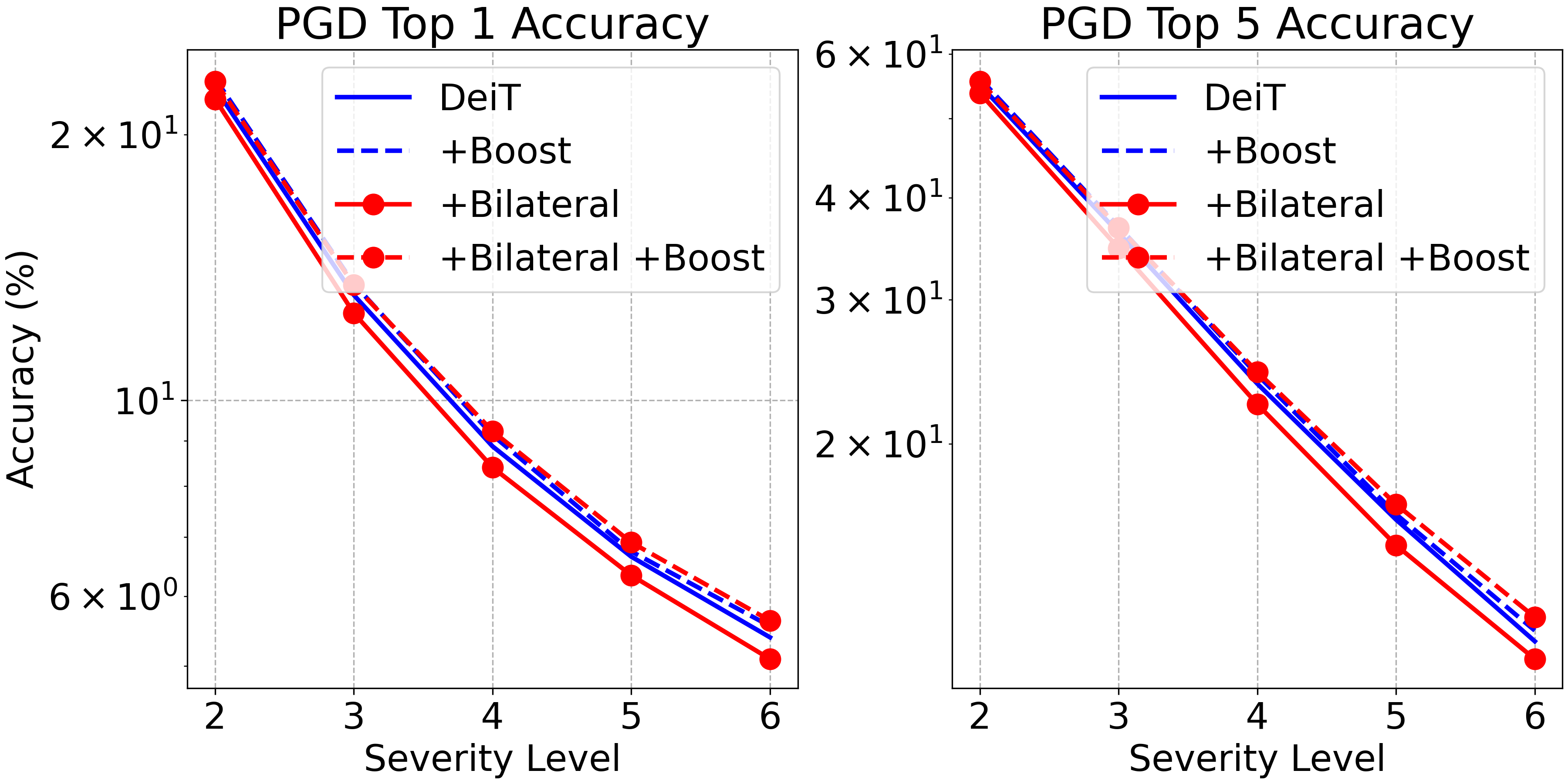}
        \label{fig:pgd}
    \end{minipage}
    \caption{Object Classification on ImageNet-1k under FGSM and PGD attacks with increasing perturbation budgets (severity level $\times$ 1/255). The figures show that models that employ boosting by GRC are consistently more robust than their counterparts without GRC. The vertical axis is in log-scale.}
    \label{fig:severe-attacks}
\end{figure*}

\begin{figure*}[t]
    \centering
    \includegraphics[width=1\linewidth]{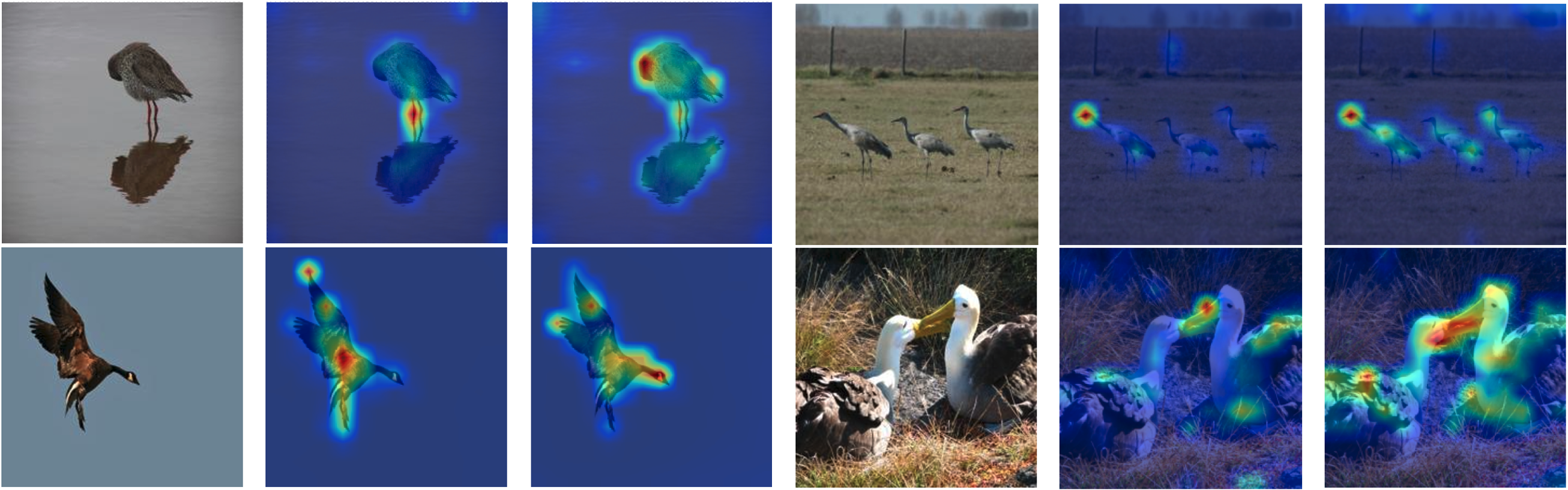}
    \caption{\textit{Left to Right:} Input, DeiT \cite{touvron2021training}, and Deit-Bilateral (ours). Attention heatmaps averaged over attention heads for DeiT \cite{touvron2021training} and DeiT-Bilateral. It can be observed that the heatmaps produced by DeiT-Bilateral captures extended meaningful regions of input samples.}
    \label{fig:attn_ptn}
\end{figure*}

\begin{figure}
    \centering
    \includegraphics[width=0.6\linewidth]{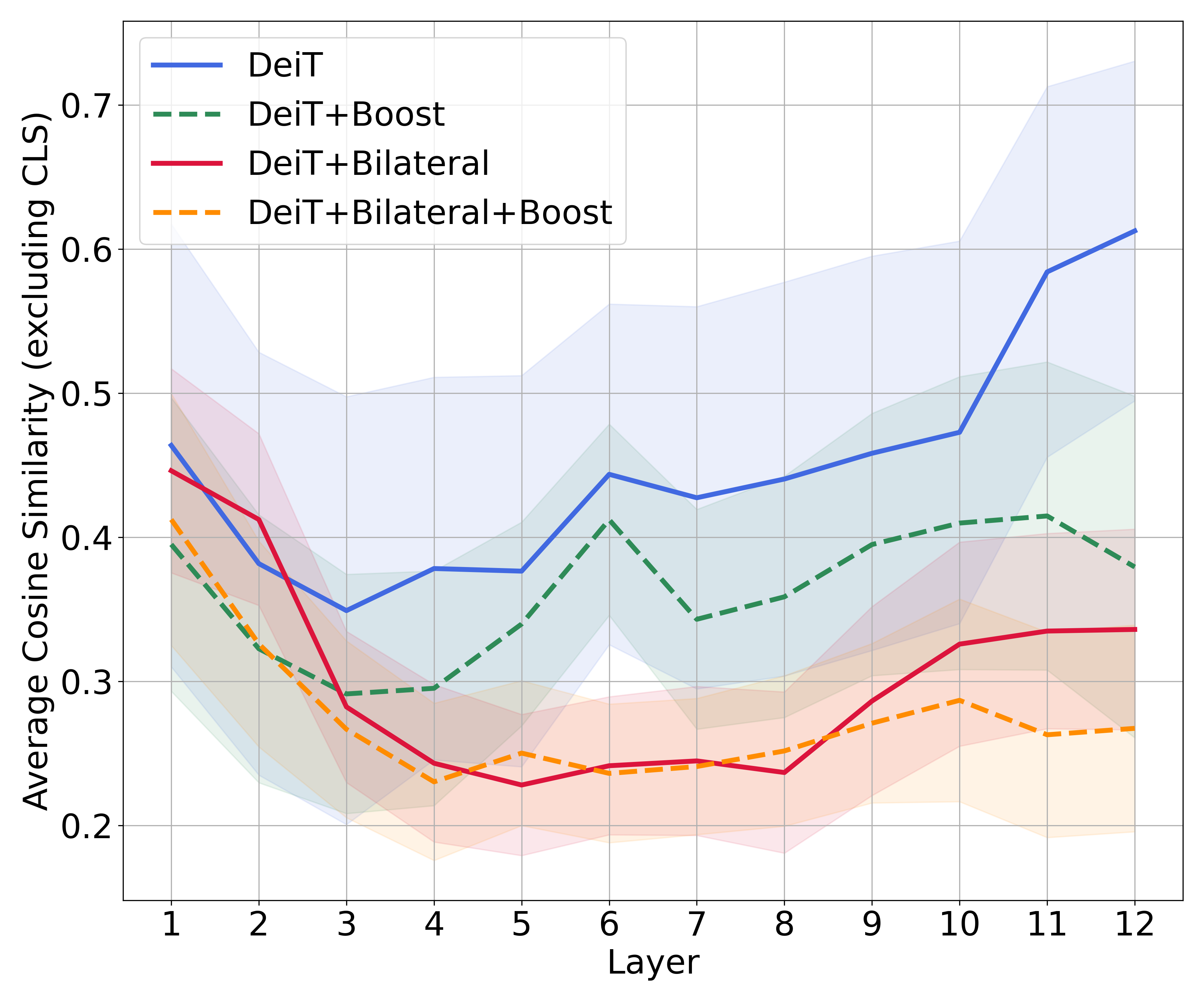}
    \caption{Average Cosine Similarity of Transformer Block Outputs Across Layers over $10^3$ random image samples. The baseline model tends to suffer from information loss by producing increasingly similar tokens over layers.}
    \label{fig:oversmoothing}
\end{figure}

\textbf{Empirical analysis.} Additionally, in Figure \ref{fig:attn_ptn}, we visualize attention heatmaps of the last layer, derived from the normalized attention weights between the class token and the image patches averaged over heads, for both DeiT with $K_{SA}$ and DeiT Bilateral with $K_{BSA}$. We note that the latter attains more coverage of semantically important objects within the image sample. As we have tested on numerous samples, we hypothesize that this behavior, when happens, is partially influenced by the presence of two additional ``rough" cross-terms (token-to-position and position-to-token) in the standard self-attention mechanism. As a result, most heads tend to focus on the region with the strongest positive signal, which can dominate these cross-term interactions, while $K_{BSA}$ is free of this distraction. Figure \ref{fig:oversmoothing}, on the other hand, shows the layer-wise average cosine similarities of the Transformer block output tokens, supporting the claim of Proposition \ref{prop:residual-signal-vanish}.


\footnotetext{Given the purpose of LRA, required attention span (RAS) serves as a proxy for how difficult a task is for Transformer-based models. The numbers are approximate copies of Figure 2 in \cite{tay2020long}.}

\section{Discussion on Framework Expressivity and Related Work}

\subsection{Attention Mechanisms and PE}

The image filtering perspective provided in this work effectively builds upon and corrects similar notions initially taken by \cite{nguyen2023mitigating} and then \cite{abdullaev2025transformer} that overlook the presence of positional information. Consequently, the framework becomes more expressive and interpretable as we shall discuss next.

Notably, the empirical observations reported in \cite{ke2021rethinking} align closely with our theoretical perspective. The authors of \cite{ke2021rethinking} identify uniform and uninformative patterns in token-to-position proximity logits, suggesting the need for independent transformations of positional vectors to mitigate their influence. Essentially, their empirical observations resembles the Bilateral Attention formulation as the following form of logits are employed:
\begin{align}
    \frac{\bm x^\top_i \bm{W}_Q^\top \bm W_{K} \bm x_j}{\sqrt{2d}} + \frac{\bm p^\top_i \bm{H}_Q^\top \bm {H}_{K} \bm p_j}{\sqrt{2d}} + b_{ij},
\end{align}
where $\bm{H}_Q$ and $\bm{H}_K$ are independent learnable projections exclusively for PE, resembling the usage of different bandwidth for BF (Section \ref{sec:nlsa}). Furthermore, Transformers with no positional encodings have also been observed to produce performance on par with absolute, relative and learnable positional encoding methods with somewhat better length generalization properties as studied by \cite{haviv2022transformer}. Our framework provides a theoretical justification for these empirical observations through the lens of NLM filtering (Section \ref{sec:nlsa}), where we connect no positional encoding to a bilateral attention with infinitesimal weights given to positional weights.

Beyond offering a unifying perspective on self-attention, our work underscores the potential for leveraging insights from classical image processing techniques to inspire novel attention mechanisms. For instance, Elliptical Attention, a recently proposed variant in \cite{nielsen2024elliptical}, computes attention scores as
\begin{align}
    \alpha_{ij} = \softmax\left(\frac{\bm q_i^\top \bm M \bm k_j}{\sqrt{d}}\right),
\end{align}
where $\bm M$ is a diagonal matrix with $M_{ii}$ representing the average function variability along the feature dimension $i$. This can be interpreted through the lens of anisotropic nonlocal means denoising techniques \cite{Maleki2013Anisotropic} and adaptive filtering methods \cite{takeda2007lark}, which construct spatially varying, non-spherical receptive fields to enhance feature preservation. Similarly, Twicing Attention \cite{abdullaev2025transformer} incorporates a residual correction technique from nonparametric regression, reminiscent of the twicing method in image processing, which our generalized framework naturally encompasses when additional computation is permitted as depicted in the following chart:

  
\bigskip

\resizebox{0.75\columnwidth}{!}{%
\begin{tikzpicture}[node distance=1cm, auto, >=Stealth]
  \node (input) [draw, rectangle, rounded corners, minimum width=1cm, minimum height=1cm] {Input $\bm X$};
  \node (filter1) [draw, rectangle, right=of input, xshift=0.5cm] {Filter};
  \node (Y) [draw, rectangle, right=of filter1, xshift=0.5cm] {$\bm Y = F(\bm X)$};
  \node (subtract) [draw, rectangle, below=of filter1, yshift=-1cm] {Residual: $\bm Y - \bm X$};
  \node (filter2) [draw, rectangle, right=of subtract, xshift=1.5cm] {Filter};
  \node (adder) [draw, rectangle, right=of Y, xshift=0.5cm] {$\bm Y + F(\bm Y - \bm X)$};
  
  \draw[->] (input) -- (filter1);
  \draw[->] (filter1) -- (Y);
  \draw[->] (input) |- (subtract);
  \draw[->] (Y) |- (subtract);
  \draw[->] (subtract) -- (filter2);
  \draw[->] (Y) -- (adder);
  \draw[->] (filter2) -- (adder);
\end{tikzpicture}%
}

\subsection{What about MLP blocks?}\label{sec:mlp}

MLPs are usually not straightforward to interpret. The interpretability of MLPs often requires them to be wide-and-sparse \cite{yang2025moeX}. Therefore, we shall discuss the sparse mixture of experts (SMoE) layer employed in most of the current state-of-the-art models \cite{shazeer2017outrageously, fedus2022switch}, conjecturing that its computation may bear a resemblance to techniques of sparse coding that are often utilized in k-SVD denoising \cite{romano2015boosting}.

In a SMoE layer, $M > 1$ MLPs $$f_j(\bm x) = \bm P_j \sigma(\bm Q_j \bm x),$$ known as \textit{experts}, are used instead of a single dense one to enhance the model capacity. Only a few of them are then activated for each input token conditionally. To be more precise, given an input token $\bm x \in \mathbb{R}^d$, the output of the SMoE layer is given by
\begin{align}
    \bm y = \sum_{j=1}^M g_j(\bm x, \bm \theta) f_j(\bm x) = \sum_{j=1}^M g_j(\bm x, \bm \theta) \bm P_j \sigma(\bm Q_j \bm x),
\end{align}
where the router scores $g_j$ for $j = 1,\dots, M$, parametrized by $\{\theta_j\}_{j=1}^M$, are computed as
\begin{align}
    g_j(\bm x, \bm \theta) = \begin{cases}
        \softmax(\bm x^\top \bm\theta_j), & \text{if } j \in \text{Top-}k(\bm x^\top \bm\theta_i) \\
        0, & \text{otherwise}.
    \end{cases}
\end{align}

Intuitively, the router functions as a means to cluster the input tokens by assigning groups of tokens to different experts by a similarity measure. This aspect of SMoE architecture has indeed been explored by a recent line of research \cite{dikkala2023on, nielsen2025tight}. We also recall that clustering is a key technique behind most image segmentation algorithms as well \cite{coleman1979image, dhanachandra2015image, yasir2020review}.

Now to explore the conjecture mentioned in the beginning of this section, we can rewrite the output in a large matrix form, which highlights the sparse and structured nature of the computation, potentially mirroring the sparse coding principles \cite{aharon2006ksvd} used in k-SVD denoising \cite{romano2015boosting}. Following \cite{liu2023sparseFFN}, let us define the intermediate output of each expert $j$ as $\bm z_j = \sigma(\bm Q_j \bm x) \in \mathbb{R}^{k'}$, where $k'$ is the dimension of the intermediate representation. The contribution of expert $j$ to the output $\bm y$ is then $g_j(\bm x, \bm \theta) \bm P_j \bm z_j$, and the total output is the sum of these contributions over all $M$ experts.

Now, define $\bm D$ as the horizontal concatenation of the output transformation matrices $\bm P_j$ for all $M$ experts $\bm D = [\bm P_1, \bm P_2, \dots, \bm P_M] \in \mathbb{R}^{d \times (M k)}$,
where each $\bm P_j \in \mathbb{R}^{d \times k'}$, so $\bm D$ has $d$ rows and $M k'$ columns. Next, construct a vector $\bm z \in \mathbb{R}^{M k'}$ by vertically stacking the scaled intermediate outputs $g_j(\bm x, \bm \theta) \bm z_j$ as $$\bm z = \begin{bmatrix}
        g_1(\bm x, \bm \theta) \bm z_1 &      g_2(\bm x, \bm \theta) \bm z_2 & \dots & g_M(\bm x, \bm \theta) \bm z_M
    \end{bmatrix}^\top.$$

    
Due to the top-$k$ selection in the router, at most $k$ of the $M$ experts have non-zero scores $g_j(\bm x, \bm \theta)$, making $\bm z$ $\frac{k}{M}$-sparse when $k \ll M$, with at most $k \times k'$ non-zero entries (since each non-zero block $g_j(\bm x, \bm \theta) \bm z_j$ has $k'$ elements) out of $M \times k'$.

With these definitions, the SMoE output can be expressed as a single matrix-vector product $$\bm y = \bm D \bm z$$ by construction. The matrix $\bm D$ can be viewed as a large ``dictionary'' of transformations, with $M k$ columns, while the sparsity of $\bm z$ ensures that only a small subset of these transformations (corresponding to the $k$ active experts) is used for each input token. We conjecture that this formulation may be related to k-SVD denoising \cite{aharon2006ksvd, romano2015boosting}, where a noisy signal is approximated using a sparse linear combination of dictionary atoms to produce a denoised output. 

\begin{table}[t]
  \scriptsize
  \centering
  \caption{Forward Pass Efficiency in terms of FLOPs (GMac)}
  \label{tab:efficiency}
  \begin{tabular}{lcc}
    \toprule
    Model & Train-time & Inference-time  \\
    \midrule
    \textit{DeiT} \cite{touvron2021training} & 1.08 & 1.08 \\
    + Boost & 1.09 & 1.09 \\
    + Bilateral & 1.33 & 1.08 \\
    \bottomrule
  \end{tabular}
\end{table}

\section{Concluding Remarks}

In this work, we developed a unifying image processing framework to better understand the feature-learning mechanism behind architectural components of Transformer-based models. In addition to addressing the frequently asked question of why positional encodings (PE) are added rather than concatenated in an investigative approach, we also explore how properly incorporating positional information enhances the accuracy and long-range capacity of the self-attention mechanism. Furthermore, we highlight the crucial role of residual connections in improving the model's fidelity to its input and, as a result, robustness against adversarial attacks.

\textbf{Future directions.} Despite the growing interest in drawing inspiration from classical image processing for designing Transformer architectures \cite{nguyen2023mitigating, abdullaev2025transformer}, several powerful techniques remain underexplored \cite{takeda2007lark, kheradmand2013general, buhlmann2003boosting, romano2017little}, and yet to be systematically transferred to Transformers or alternative models. Similarly, fast approximations of the bilateral filter \cite{Paris2006}, which enable efficient edge-preserving smoothing in real-time applications, could have been considered as a foundation for designing computationally efficient self-attention mechanisms or linear recurrent models.

\textbf{Limitations.} A specific limitation of Bilateral Attention proposed in this work is that it introduces an additional training-time computation complexity (note it does not pose any additional inference cost since positional embeddings are input-independent and can be precomputed after training, see Table \ref{tab:efficiency}), which is undesirable under specific resource restricted scenarios. Moreover, a room of ambiguity still remains when it comes to understanding MLP blocks (if possible) other than the attempts in the previous section, which is an interesting open question for further exploration.

\appendix

\section{Technical Proofs}

\subsection*{Proof of Theorem \ref{thm:self-attention-denoising}}

The proof is similar to the derivation of Eqn.~\ref{eq:weighted-averaging-soln} but with vector valued intensity values. The signal $u(\bm{p}) \in \mathbb{R}^d$ at position $\bm{p}$ is estimated using a (nonparametric) 
point estimation framework. Specifically, this involves solving a weighted 
least squares problem:
\begin{align}
    \hat{u}(\bm{p}_i) = \arg \min_{u(\bm{p}_i)} \mathcal{J}_K(u(\bm p_i)),
\end{align}
where Euclidean distance metric is used as follows:
\begin{align}
    \mathcal{J}_K(u(\bm p_i)) = \sum_{j=1}^n \|\bm y_j - u(\bm{p}_i)\|^2 K(\bm{p}_i, \bm{p}_j, \bm y_i, \bm y_j),
\end{align}
and the weight function $K(\cdot)$ is a symmetric and positive kernel
with respect to the indices $i$ and $j$, quantifying the ``similarity" between samples $\bm y_i$ and $\bm y_j$, located at positions $\bm{p}_i$ and $\bm{p}_j$, respectively. The stationary point condition for an optimal $\hat u$ is given by
\begin{align}
    \nabla_{u}\mathcal{J}_K \big|_{\hat u} = \sum_{j=1}^{n} -2[\bm y_j - \hat u(\bm p_i)] K(\bm{p}_i, \bm{p}_j, \bm y_i, \bm y_j) = 0.
\end{align}
Then, for any given kernel function $K(\cdot)$, the solution takes the following form:
\begin{align}
    \hat{u}(\bm{p}_i) = \frac{\sum_{j=1}^{n} \bm y_j K(\bm{p}_i, \bm{p}_j, \bm y_i, \bm y_j)}{\sum_{j=1}^{n} K(\bm{p}_i, \bm{p}_j, \bm y_i, \bm y_j)}.
\end{align}
The proof is completed by plugging in the kernel $K_{SA}$. \qed

\subsection*{Proof of Theorem \ref{thm:softmax-unstability}}\label{app:proof-asymp}

First, note that the Lipschitz continuity of softmax gives
\begin{align}
    \| \softmax(\bm c + \bm\eta) - \softmax(\bm c) \| \le L \| \bm \eta \|.
\end{align}
It remains to bound $\mathbb{E}\| \bm \eta \|$. Without loss of generality, assume that $\sigma = 1$ so that $\mathbb{E}[\eta_i^2] = 1$ for all $i \in [N]$. 
 
To estimate the remainder term, denote $\xi := \|\bm \eta\|^2/N$. Note that
\begin{align}
    \mathbb{E}[\xi] = \mathbb{E}\left[\frac{1}{N}\sum_{i=1}^N \eta_i^2\right] = \frac{1}{N}\sum_{i=1}^N \mathbb{E}\left[\eta_i^2\right] = 1,
\end{align}
and $\text{Var}(\xi) = 1/N$. This implies that $\xi$ is usually very close to $1$. In fact, for any $\varepsilon > 0$, Chebyshev's inequality states that
\begin{align}
    \Pr(\left| \xi - 1 \right| \le \varepsilon) \ge 1 - \frac{1}{N \varepsilon^2}.
\end{align}
This observation hints $\|\bm \eta \| \sim \sqrt{N}$. To make it rigorous, we look at the following expansion:
\begin{align}
    \sqrt{\xi} = \sqrt{1 + (\xi - 1)} = 1 + \frac{\xi - 1}{2} - \frac{(\xi - 1)^2}{8} + o\left((\xi - 1)^2\right). \nonumber
\end{align}
Now since $\sqrt{\xi}$ is a concave function of $\xi$, it is below its tangent, and also there exists an absolute constant $\gamma >0$ such that
\begin{align}
    1 + \frac{\xi - 1}{2} - \frac{(\xi - 1)^2}{\gamma} \le \sqrt{\xi} \le 1 + \frac{\xi - 1}{2}.
\end{align}
It is easy to verify that taking any $\gamma \in (0, 2]$ is sufficient. Now taking expectation throughout the inequalities and noticing that $\mathbb{E}[\xi - 1] = 0$ and $\mathbb{E}[(\xi - 1)^2] = \text{Var}(\xi) =  1/N$, we arrive at
\begin{align}
    1 - \frac{1}{2N} \le \mathbb{E}\left[\sqrt{\xi}\right] \le 1 \implies \left|\mathbb{E}\|\bm\eta\| - \sqrt{N}\right| \le \frac{1}{2\sqrt{N}}. \nonumber
\end{align}
Note that since the inequalities applied after the Lipschitzness of softmax are asymptotically tight in $N$, the final bound is also asymptotically tight. This completes the proof. \qed



\subsection{Proof of Theorem \ref{thm:robustness}}\label{app:robustness}

The difference at each layer satisfies:
\begin{align}
    \| \bm{Y}_{\ell+1} - \bm{Y}_{\ell+1}' \| &= \| f_\ell(\bm{Y}_\ell) - f_\ell(\bm{Y}_\ell') + \bm{Y}_\ell - \bm{Y}_\ell' \| \notag \\
    &\leq L_\ell \| \bm{Y}_\ell - \bm{Y}_\ell' \| + \| \bm{Y}_\ell - \bm{Y}_\ell' \| \notag \\
    &= (L_\ell + 1) \| \bm{Y}_\ell - \bm{Y}_\ell' \|.
\end{align}
Thus, the recurrence becomes:
\begin{align}
    K_{\ell+1} = (L_\ell + 1) K_\ell, \quad K_1 = L_0 + 1.
\end{align}
Assuming $L_\ell = L$ for all $\ell$, we get $K_{\text{RC}} \asymp (L + 1)^n$.

Now consider the modified connection. The difference at each layer is $\bm{Y}_{\ell+1} - \bm{Y}_{\ell+1}' = f_\ell(\bm{Y}_\ell) - f_\ell(\bm{Y}_\ell') + t (\bm{Y}_0 - \bm{Y}_0') + (1 - t)(\bm{Y}_\ell - \bm{Y}_\ell')$ and, thus,
\begin{align}
    \| \bm{Y}_{\ell+1} - \bm{Y}_{\ell+1}' \| &\leq L_\ell \| \bm{Y}_\ell - \bm{Y}_\ell' \| + t \| \bm{Y}_0 - \bm{Y}_0' \|+ (1 - t) \| \bm{Y}_\ell - \bm{Y}_\ell' \| \nonumber \\
    &= [L_\ell + (1 - t)] \| \bm{Y}_\ell - \bm{Y}_\ell' \| + t \| \bm{Y}_0 - \bm{Y}_0' \|. \nonumber
\end{align}
Using the induction hypothesis $\| \bm{Y}_\ell - \bm{Y}_\ell' \| \leq K_\ell \| \bm{Y}_0 - \bm{Y}_0' \|$, we get:
\begin{align}
    K_{\ell+1} = [L_\ell + (1 - t)] K_\ell + t, \quad K_1 = L_0 + 1.
\end{align}
Assuming $L_\ell = L$, define $a = L + (1 - t)$, $b = t$, and solve the recurrence $K_{\ell+1} = a K_\ell + b$. The closed-form solution for $a \neq 1$ is then given by
\begin{align}
    K_n = (K_1 - \tfrac{b}{a - 1}) a^{n-1} + \tfrac{b}{a - 1}.
\end{align}
Thus, the dominant term for large $n$ is $K_{\text{GRC}} \asymp [L + (1 - t)]^n$.
Observe that for any $t \in (0,1]$, we have $L + (1 - t) < L + 1$, implying slower exponential growth in the modified connection as $$\frac{K^{\text{GRC}}_n}{K^{\text{RC}}_n} \asymp \left( 1 - \frac{t}{L + 1} \right)^n,$$ and a better robustness. \qed

\section*{Acknowledgments}
We would like to acknowledge the assistance of volunteers in putting
together this example manuscript and supplement.

\bibliographystyle{siamplain}
\bibliography{references}

\end{document}